\documentclass[journal]{IEEEtran}

\usepackage[latin1]{inputenc}
\usepackage{amsmath}
\usepackage{graphicx}
\usepackage{listings}
\usepackage[colorlinks,linkcolor=black,citecolor=black,urlcolor=black]{hyperref}
\usepackage{siunitx}
\usepackage{fixltx2e}
\usepackage{multirow}
\usepackage{threeparttable}
\usepackage{tabularx}
\usepackage{booktabs}
\usepackage{units}
\usepackage{xcolor}
\usepackage[normalem]{ulem}
\newif\ifshowToDos

\usepackage{filecontents}
\usepackage[noadjust]{cite}

\showToDosfalse

\ifshowToDos
  \usepackage{todonotes}
  \paperwidth=\dimexpr \paperwidth + 8cm\relax
  \oddsidemargin=\dimexpr\oddsidemargin + 4cm\relax
  \evensidemargin=\dimexpr\evensidemargin + 4cm\relax
  \marginparwidth=\dimexpr \marginparwidth + 4cm\relax
\else
  \usepackage[disable]{todonotes}
\fi

\renewcommand{\d}{d}
\newcommand{\ve}[1]{\boldsymbol{#1}}

\newcommand{\expect}[2][]{\left\langle\, #2\, \right\rangle_{#1} }

\definecolor{gray}{gray}{0.2}

\newcommand{\wiener}{\mathcal{W}}

\newcommand{\ddthetai}{\frac{\partial}{\partial \theta_i}}

\newcommand{\syn}[1]{\text{\textsc{syn}}_{#1}}
\newcommand{\prei}{\text{\tiny{\textsc{pre}}}_{i}}
\newcommand{\posti}{\text{\tiny{\textsc{post}}}_{i}}
\newcommand{\preidef}{\text{\textsc{pre}}_{i}}
\newcommand{\postidef}{\text{\textsc{post}}_{i}}

\newcommand{\zk}{z_{\posti}}
\newcommand{\zkt}{\zk(t)}

\newcommand{\fk}{f_{\posti}}
\newcommand{\fkt}{\fk(t)}

\newcommand{\rt}{r(t)}
\newcommand{\rtau}{r(\tau)}
\newcommand{\rhatt}{\hat{r}(t)}

\newcommand{\hz}{y}

\newcommand{\bth}{\ve \theta}

\newcommand{\thi}{\theta_i(t)}

\newcommand{\ps}[1]{p_{\mathcal{S}}\left({#1}\right)}

\begin{document}
\bstctlcite{IEEEexample:BSTcontrol}
\title{Efficient Reward-Based Structural Plasticity on a SpiNNaker 2 Prototype}

\author{
  Yexin Yan$^1$, David Kappel$^{1,2,4}$, Felix Neum\"arker$^1$, Johannes Partzsch$^1$,\\
  Bernhard Vogginger$^1$, Sebastian H\"oppner$^1$, Steve Furber$^3$, Wolfgang Maass$^2$, Robert Legenstein$^2$, Christian Mayr$^1$\\
  $^1$ Technische Universit\"at Dresden, Chair of Highly Parallel\\
  VLSI Systems and Neuromorphic Circuits, Dresden, Germany,\\
  $^2$ Technische Universit\"at Graz, Institute for Theoretical Computer Science, Graz, Austria,\\
  $^3$ University of Manchester, School of Computer Science, Manchester, UK,\\
 $^4$ Bernstein Center for Computational Neuroscience, III Physikalisches Institut-Biophysik, Georg-August Universit\"at, G\"ottingen, Germany\\
  Corresponding author: Yexin Yan, Email: yexin.yan@tu-dresden.de
}
© 2019 IEEE.  Personal use of this material is permitted.  Permission from IEEE must be obtained for all other uses, in any current or future media, including reprinting/republishing this material for advertising or promotional purposes, creating new collective works, for resale or redistribution to servers or lists, or reuse of any copyrighted component of this work in other works.
\clearpage
\maketitle

\begin{abstract}
  Advances in neuroscience uncover the mechanisms employed by the brain to efficiently solve complex learning tasks with very limited resources.
  However, the efficiency is often lost when one tries to port these findings to a silicon substrate, since brain-inspired algorithms often make extensive use of complex functions such as random number generators, that are expensive to compute on standard general purpose hardware.
  The prototype chip of the 2nd generation SpiNNaker system is designed to overcome this problem.
  Low-power ARM processors equipped with a random number generator and an exponential function accelerator enable the efficient execution of brain-inspired algorithms.
  We implement the recently introduced reward-based synaptic sampling model that employs structural plasticity to learn a function or task.
  The numerical simulation of the model requires to update the synapse variables in each time step including an explorative random term.
  To the best of our knowledge, this is the most complex synapse model implemented so far on the SpiNNaker system.
  By making efficient use of the hardware accelerators and numerical optimizations the computation time of one plasticity update is reduced by a factor of 2.
  This, combined with fitting the model into to the local SRAM, leads to 62\% energy reduction compared to the case without accelerators and the use of external DRAM.
  The model implementation is integrated into the SpiNNaker software framework allowing for scalability onto larger systems.
  The hardware-software system presented in this work paves the way for power-efficient mobile and biomedical applications with biologically plausible brain-inspired algorithms.
  
\end{abstract}

\begin{IEEEkeywords}
  SpiNNaker chip, random number generator, exponential function accelerator, neuromorphic computing, Bayesian reinforcement learning, synaptic sampling, structural plasticity
\end{IEEEkeywords}

\section{Introduction}
\label{sec:intro}

Neurophysiological data suggest that brain networks are sparsely connected, highly dynamic and noisy \cite{Faisal2008,Clarke2012}. A single neuron is only connected to a fraction of potential postsynaptic partners and this sparse connectivity changes even in the adult brain on the timescale of hours to days \cite{Holtmaat2005,Rumpel2016}. The dynamics that underlies the process of synaptic rewiring was found to be dominated by noise \cite{Dvorkin2016}.
It has been further suggested that the permanently ongoing dynamics of synapses lead to a random walk that is well described by a stochastic drift-diffusion process, that gives rise to a stationary distribution over synaptic strengths. Therefore, synapses are permanently changing and randomly rewiring while the overall statistics of the connectivity remains stable \cite{Rokni07,YasumatsuETAL:08,LoewensteinETAL:11,Statman14}.
Theoretical considerations suggest that the brain is not suppressing these noise sources since they can be exploited as a computational resource to drive exploration of parameter spaces, and several models have been proposed to capture this feature of brain circuits (see \cite{McDonnell2011} and \cite{Maass2014} for reviews).

The \emph{synaptic sampling} model that has been proposed in \cite{Kappel2015,kap15} employs this approach for rewiring and synaptic plasticity. The noisy learning rules drive a sampling process which mimics the drift-diffusion dynamics of synapses in the brain. Although the network is permanently rewired, this process provably leads to a stationary distribution of the connectivity. This distribution over the network connectivity can be shaped by reward signals, to incorporate reinforcement learning, and can be constrained to enforce sparsity \cite{kap17}. The synaptic sampling model reproduces a number of experimental observations, such as the dynamics of synaptic decay under stimulus deprivation or the long-tailed distribution over synaptic weights \cite{Kappel2015,kap17}. Furthermore, when equipped with standard error back-propagation this method was found to perform on a par with classical fully connected machine learning networks, at a fraction of the memory requirement \cite{bel17}.

However, the gain in efficiency of biology-inspired algorithms such as synaptic sampling can often not be fully realized on either dedicated neuromorphic hardware or standard digital compute hardware, since these models require complex operations such as random number generation or exponential functions. The former hardware usually has very narrowly configurable plasticity functions unsuitable for this kind of exploration  \cite{indiveri2015neuromorphic,noack2015switched,du2015single,levi2018}. Thus, synaptic weights that experience complex plasticity functions are usually precomputed in software and then run statically on mixed-signal \cite{Schmitt2017,petrovici2017pattern} or on digital neuromorphic hardware \cite{merolla2014}. On the other hand, standard digital compute hardware is in principle flexible enough, but the functions required by the plasticity models are very expensive to compute on standard hardware which significantly narrows down the gain in efficiency. Despite recent efforts to simulate spiking neural networks on GPUs \cite{knight18}, there is, to the best of our knowledge, no hardware support available for random number generation, especially true random number generation, and exponential function in GPUs.
A common workaround on digital hardware is to store massive amount of random numbers and look-up tables for the exponential function before the simulation starts \cite{vog15}. This reduces computation time at the cost of increasing the requirements for the already limited memory of embedded applications. The 2nd generation SpiNNaker system strives to break the trade-off between computation time and memory by employing dedicated hardware components for these time- (and energy-)consuming operations. Standard ARM processors are augmented with hardware accelerators for random numbers \cite{fel16} and exponential functions \cite{par17}. 
We show that this allows us to implement complex learning algorithms in a compact, power efficient package. 
In addition, by fitting the model into the local SRAM, DRAM can be switched off, further reducing the power consumption.
This potentially offers a new compute substrate especially for mobile and biomedical applications such as neural implants that are strictly limited by the power budget, computation speed and memory capacity of the silicon chip on which they are executed.

In this article we present the main features of the prototype chip of the 2nd generation SpiNNaker system in detail and showcase the benefits of the architecture for experiments on reward-based synaptic sampling \cite{kap17}. We show that the architecture allows us to exploit the advantage of the synaptic sampling algorithm. The model is efficiently implemented thanks to the hardware accelerators, the software optimizations and the floating point unit available in ARM M4F. We show a speedup of more than 2 due to the use of hardware accelerators. Our hardware-software system optimizes the implementation of reward-based synaptic sampling with respect to the memory footprint, computation and power and energy consumption. We built a scalable distributed real-time online learning system and demonstrate its usability in a closed-loop reinforcement learning task.
Furthermore, we study a modified rewiring scheme called \emph{random reallocation} that recycles the memory of synapses by immediately reconnecting them to a new post-synaptic target. We show that this more efficient version of synaptic sampling also leads to faster learning.


In Section~\ref{sec:hardware} we give an overview of the prototype chip, focusing on the random number generator and the exponential function accelerator. Section~\ref{sec:synapse_model} shows the reward-based synaptic sampling model implemented in this work. Section~\ref{sec:software} presents the software implementation and experimental results are presented in Section~\ref{sec:results}.





\section{Hardware}
\label{sec:hardware}

\subsection{System Overview}\label{sys_ove}

\begin{figure}[htb]
  \centering
  \includegraphics[width=0.48\textwidth]{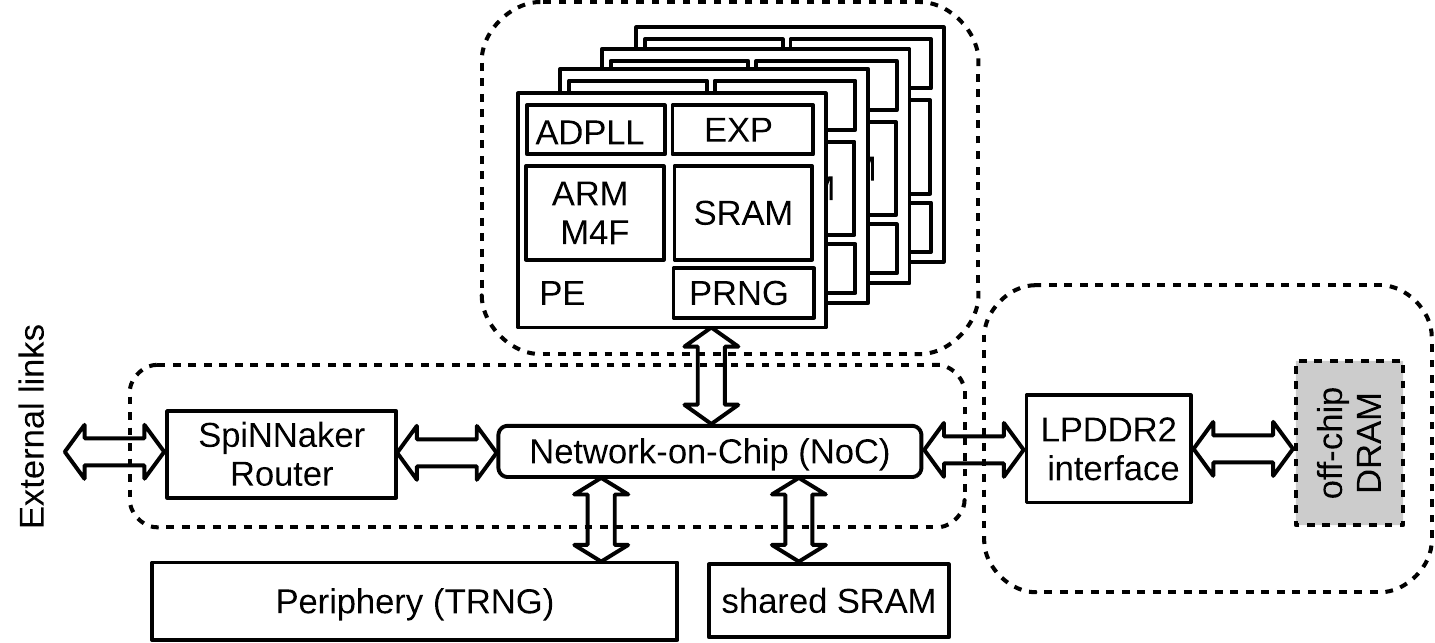}
  \caption{\label{fig:overview}Overview of the SpiNNaker 2 prototype including 4 processing elements (PE) with ARM core, power management controller (PMC) and exponential function accelerator (EXP), True Random Number Generator (TRNG), Network-on-Chip (NoC), SpiNNaker router, shared on-chip SRAM (not used in this work) and off-chip DRAM}
\end{figure}

SpiNNaker \cite{fur14} is a digital neuromorphic hardware system based on low-power ARM processors built for the real-time simulation of spiking neural networks (SNNs).
On the basis of the first-generation SpiNNaker architecture and our previous work in power efficient multi-processor systems on chip \cite{haas2016mpsoc,haas2017heterogeneous}, the second generation SpiNNaker system (SpiNNaker~2) is currently being developed in the Human Brain Project \cite{amunts2016human}.
By employing a state-of-the art CMOS technology and advanced features such as per-core power management, more processors can be integrated per chip at significantly increased energy-efficiency.
In this article we use the first SpiNNaker~2 prototype chip, with architecture as shown in Fig.~\ref{fig:overview}.
Table~\ref{tab:hwspinn} provides a brief summary of the new hardware features which are relevant for this work, in contrast to the first generation SpiNNaker~\cite{pai13} system.
Furthermore, the table includes an outlook on the final SpiNNaker~2 chip (tape-out 2020).

\begin{table*}
  \begin{center}
    \caption{Comparison of SpiNNaker 1 and SpiNNaker 2}\label{tab:hwspinn}
    \begin{tabular}{lccc}\toprule
      & SpiNNaker 1     & SpiNNaker 2 Prototype & SpiNNaker 2 \\
                          &                 & (used in this work)   & (current plan, cf. \cite{hoppner2018spinnaker2}) \\ \midrule
        Microarchitecture & ARMv5TE         & ARMv7-M               & ARMv7-M       \\
        Max. Clock Frequency & 200 MHz         & 500 MHz               & 500 MHz \\
        Floating Point    & ---             & single precision      & single precision \\
        HW Accelerators   & ---             & EXP, PRNG, TRNG       & EXP, LOG, PRNG, TRNG \\
        Technology node  & 130 nm           & 28 nm                 & 22 nm \\
        ARM cores / chip  & 18               & 4                     & 144 \\ \bottomrule
    \end{tabular}
  \end{center}
\end{table*}

\begin{figure}[htb]
  \centering
  \includegraphics[width=0.4\textwidth]{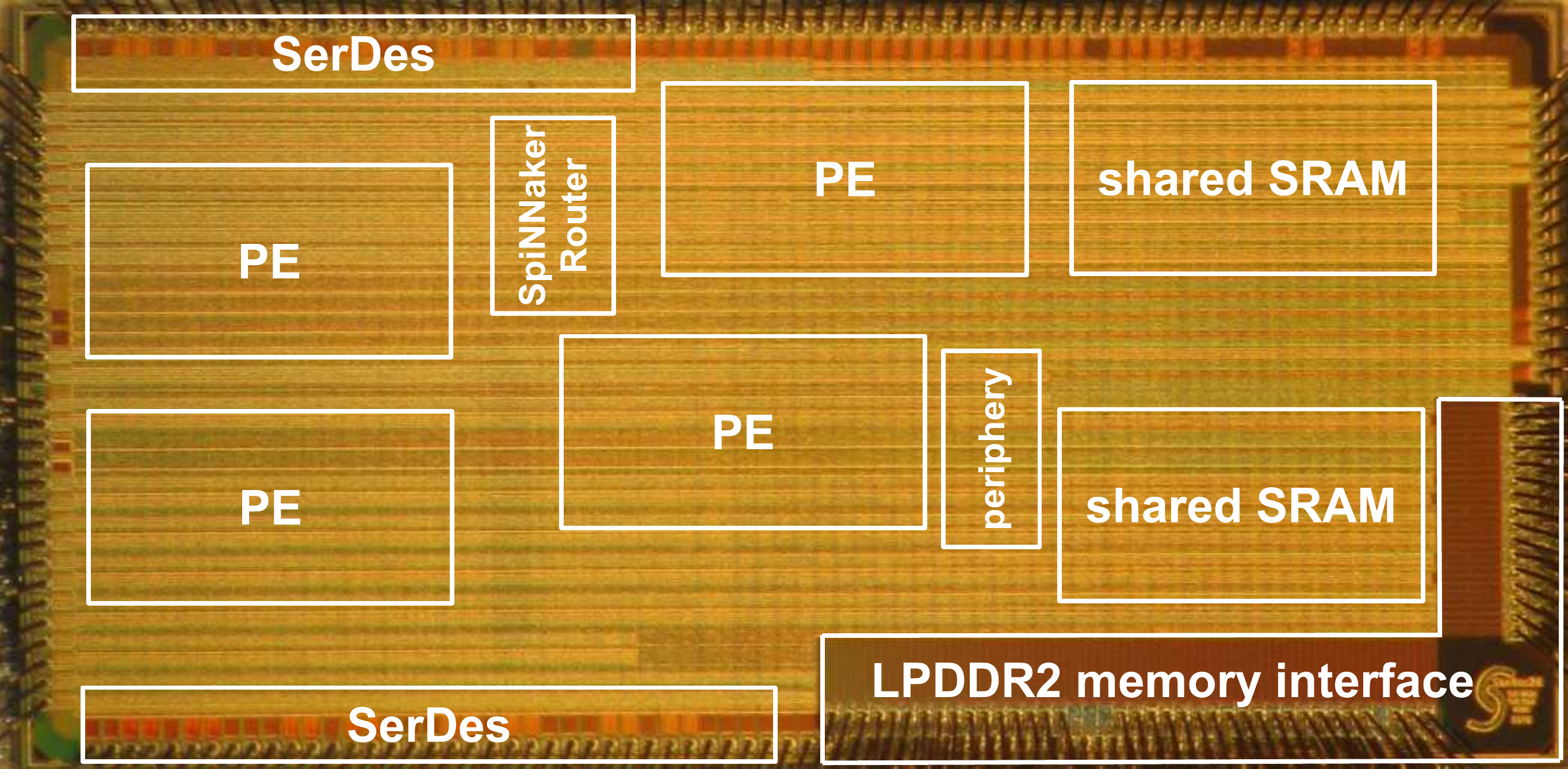}
  \caption{\label{fig:photo} Photo of the prototype chip fabricated in 28 nm technology, with the location of the building blocks~\cite{Hoeppner2017}.}
\end{figure}

The processing element (PE) is based on an ARM M4F processor core with 128 KB local SRAM, an exponential function accelerator~\cite{par17}, neuromorphic power management~\cite{Hoeppner2017} and a hardware pseudo random number generator (PRNG).
The SpiNNaker router~\cite{nav15} handles on-chip and off-chip spike communication.
Furthermore the chip provides a dedicated true random number generator (TRNG). The various components are interconnected via Network-on-Chip (NoC).
The chip has been fabricated in 28 nm SLP CMOS technology by \textsc{Globalfoundries} (Fig.~\ref{fig:photo}).

The next two Sections~(\ref{sec:ran_num},~\ref{sec:exp_fun}) will give an introduction of the hardware accelerators, i.e., the random number generator and the exponential function accelerator.

\subsection{Random Number Generator}\label{sec:ran_num}

The hardware PRNG is a specific implementation of Marsaglia's KISS~\cite{mar03} random number generator.
The generated sequence depends only on the initial seed.
The provided 32-Bit integer values are uniform distributed and accessible within a delay of one clock cycle.
An equivalent software implementation takes $35$ clock cycles \footnote{All clock cycle numbers in this paper are measured on the ARM core of the prototype chip}.
The model in this work uses uniform distributed floating-point numbers in the range from $0$ to $1$.
Therefore, the conversion to floating point and the range scaling adds another $7$~clock cycles, resulting in 42 clock cycles in total.
\par
The main advantage of a PRNG over a TRNG is the reproducibility, which simplifies debugging.
However, due to the properties of a PRNG not all effects of the randomness might be seen, since the entropy of the sequence is reduced to the seed of the generator.
In order to facilitate to run an experiment with different random inputs and a higher entropy, the prototype offers the possibility to scramble the seed of the PRNG
with a value generated by the TRNG.
From a software point of view just the initial configuration differs and no further changes on the code are necessary.
The entropy source of the TRNG is the jitter of the different clock-generators of the chip \cite{hoppner2013fast}. In conventional clock generators, this unwanted noise would be cancelled by the control loop \cite{eisenreich2009novel}. However, in this case the noise provides us with an entropy source at minimal cost in terms of power and area, since the clock-generators have to run anyway, for the PE itself as well as for the SpiNNaker links.  The principle is described in detail in~\cite{fel16} and has been submitted as a patent~\cite{trngpat}. The entropy of each single clock-generator is combined as true random bus which is sampled by the PRNG in order to realize the scrambling.

\subsection{Exponential Function Accelerator}\label{sec:exp_fun}

The exponential function accelerator calculates an exponential function with the signed fixed-point s16.15 data type. In the implementation, the operand is divided into three parts:

\begin{equation}
 y = \mathrm{exp}(x) = \underbrace{\mathrm{exp}(n)}_{f_\mathrm{int}(n)} \cdot \underbrace{\mathrm{exp}(p)}_{f_\mathrm{frac}(p)}
                       \cdot \underbrace{\mathrm{exp}(q)}_{f_\mathrm{poly}(q)}
 \;\mbox{with}\;x=n+p+q ,
 \label{eq:exp_split}
\end{equation}
where \(n\) is the integer part, \(p\) and \(q\) are the upper and lower fractional parts, respectively. $f_\mathrm{int}(n)$ and $f_\mathrm{frac}(p)$ are calculated with two separate look-up tables (LUTs), and $f_\mathrm{poly}(q)$ is a polynomial. The split into two separate LUTs considerably reduces the memory size and thus the silicon area compared to one combined LUT, by taking advantage of the properties of the exponential function. The split of the evaluation of the fractional part into a LUT and a polynomial reduces the computational complexity of the polynomial with minimum memory overhead. The overall implementation achieves single-LSB precision in the employed fixed-point format \cite{par17}. The exponential accelerator is included in each PE, and makes up for approx.~2\% of the silicon area of each PE. The look-up and the polynomial calculation are parallelized, resulting in a latency of four clock cycles for each exponential function. Writing the operand to the accelerator and reading the result from it via the AHB bus adds additional two clock cycles, resulting in 6 clock cycles in total. In pipelined operation the processor writes one operand in one clock cycle and reads the result of a previous exponential function in another clock cycle, resulting in two clock cycles per exponential function \cite{par17}.

\section{Spiking network model}
\label{sec:synapse_model}

To demonstrate the performance gain of the SpiNNaker~2 hardware for simulations of spiking neural networks, we implemented the \emph{synaptic sampling} model introduced in~\cite{kap17}.
In this section we briefly review this model for stochastic synaptic plasticity and rewiring.
The model combines insights from experimental results on synaptic rewiring in the brain with a model for online reward maximization through policy gradient (see Section \ref{sec:reward-based-sampling} for details). The network has a large number of \emph{potential synaptic connections} only a fraction of which is functional at any moment in time, whereas most others are non-functional (disconnected). The network connectivity is permanently modified through rewiring. Synaptic weight changes and rewiring are guided by stochastic learning rules that probe different network configurations. Hence, synaptic sampling, other than usually considered deterministic learning rules that converge to some (local) optimum of parameters, in our framework learning converges to a target distribution $p^*(\bth)$ over synaptic parameters $\bth$. The learning rules are designed in such a way that maxima of the distribution $p^*(\bth)$ coincide with maxima of the expected reward. We first summarize the general synaptic sampling framework in Section \ref{sec:synapse-model} and \ref{sec:neuron-model} and then provide additional details to its application to reinforcement learning in Section \ref{sec:reward-based-sampling}.
All parameter values are summarized in Table~\ref{tab:params}.
In Section \ref{sec:ran_rew} we discuss \emph{random reallocation of synapses}, a modified rewiring scheme that is more memory efficient.

\subsection{Synapse model}
\label{sec:synapse-model}
In our model for synaptic rewiring we consider a neural network scaffold with a large number of potential synaptic connections between  neurons. For each functional synaptic connection, we introduce a real-valued parameter $\theta_i$ that determines the strength $w_i$ of connection $i$ through the exponential mapping
\begin{equation}
w_{i} \;=\; \exp( \theta_i - \theta_0 )
\label{eq:thetamap}
\end{equation}
with a positive offset parameter $\theta_0$ that scales the minimum strength of synaptic connections. 
The mapping in Eq.~\eqref{eq:thetamap} accounts for the experimentally found multiplicative synaptic dynamics in the cortex (c.f.~\cite{HoltmaatETAL:06,YasumatsuETAL:08,LoewensteinETAL:11}, see~\cite{kap17} for details). For simplicity we assume that only excitatory connections (with $w_i \geq 0$) are plastic, but the model can be easily generalized to inhibitory synapses.

The functional goal of network learning is determined by the dynamics of the synaptic parameters $\theta_i$.
It was shown in~\cite{kap17} that for some target distribution $p^{*}(\bth)$ over synaptic parameters with partial derivative $\left.
\ddthetai \log p^{*}(\bth) \right|_{t}$ of the log-distribution with respect to parameter $\theta_i$ evaluated at time $t$, the stochastic drift-diffusion processes
\begin{equation}
d \theta_i(t) \;=\; \beta \, \left.
\ddthetai \,\log p^*(\bth) \right|_{t} \, dt  \;+ \; \sqrt{2 \beta T} \, d \wiener_{i} (t)
\label{eq:sde-reduced}
\end{equation}
give rise to a stationary distribution over $\bth$ that is proportional to $p^{*}{(\bth)}^{\frac{1}{T}}$.
In Eq.~\eqref{eq:sde-reduced} $\beta$ plays the role of a learning rate and $d \wiener_i$ are stochastic increments and decrements of Wiener processes, which are scaled by the \emph{temperature parameter} $T$.

This result suggests that a rule for reward-based synaptic plasticity should be designed in a way that $p^{*}(\bth)$ has most of its mass on highly rewarded parameter vectors $\bth$. We use target distributions $p^{*}(\bth)$ of the form $p^*(\bth) \;\propto\; p_S(\bth) \, \times \, \mathcal{V}(\bth)$ where $\propto$ denotes proportionality up to a positive normalizing constant. $p_S(\bth)$ can encode structural priors of the network scaffold, e.g. to enforce sparsity. This happens when $p_S(\bth)$ has most of its mass near $\boldsymbol{0}$. In our experiments we have used a Gaussian distribution with mean $\mu$ and variance $\sigma^2$ for the prior $\ps{\bth}$, such that $\ddthetai \,\log \ps{\bth} = \frac{1}{\sigma^2}\left( \mu - \thi \right)$.

The function $\mathcal{V}(\bth)$ denotes the expected discounted reward associated with a given parameter vector $\bth$. In Section~\ref{sec:reward-based-sampling} we will discuss in detail how the term $\ddthetai \log \mathcal{V}(\bth)$ can be computed using reward-modulated plasticity rules.

Synaptic rewiring is included in this model by interpreting each synapse $i$ for which $\theta_i \leq 0$ as disconnected. To reconnect synapses we tested two approaches. In the first approach we continued to simulate the dynamics of the prior distribution, i.e. a process of the form \eqref{eq:sde-reduced} with $p^*(\bth) = p_S(\bth)$ until the synapse reconnects ($\theta_i>0$). This is the algorithm that was proposed in \cite{kap17}. In Section~\ref{sec:ran_rew} we introduce another approach for rewiring called \emph{random reallocation of synapses} that makes more effective use of memory resources. The two approaches are compared in the results below.

\subsection{Neuron model}
\label{sec:neuron-model}

We considered a general network of $K$ stochastic spiking neurons and we denote the output spike train of a neuron $k$ by $z_k(t)$, defined as the sum of Dirac delta pulses positioned at the spike times $t_k^{(1)}, t_k^{(2)}, \dots$, i.e., $z_k(t) = \sum_l \delta(t-t_k^{(l)})$.
We denote by $\preidef$ and $\postidef$ the index of the pre- and postsynaptic neuron of synapse $i$,
respectively, which unambiguously specifies the connectivity in the network.
Further, we define $\syn{k}$ to be the index set of synapses that project to neuron $k$.
Note that this indexing scheme allows us to include  multiple (potential) synaptic connections between a given pair of neurons. In all simulations we allow multiple synapses between neuron pairs.

Network neurons were modeled by a standard stochastic variant of the spike response model~\cite{GerstnerETAL:14}.
We denote by $w_{i}(t)$ the synaptic efficacy of the $i$-th synapse in the network at time~$t$, determined by Eq.~\eqref{eq:thetamap}. The membrane potential of neuron $k$ at time $t$ is then given by

\begin{equation}
  u_k(t) \;=\; \sum_{i \,\in\, \syn{k}} \hz_{\prei}(t)\, w_{i}(t) \;+\; \vartheta_k(t) \; , \label{eq:membrane-potential}
\end{equation}
where $\vartheta_k(t)$ denotes the slowly adapting bias potential of neuron $k$, and $\hz_{\prei}(t)$ denotes the trace of the (unweighted) postsynaptic potentials (PSPs) that neuron $\preidef$ leaves in its postsynaptic synapses at time $t$.
It is defined as $\hz_{\prei}(t) = z_{\prei}(t) \ast \epsilon(t)$ given by spike trains filtered with a PSP kernel of the form $\epsilon(t) = \Theta(t) \, \frac{\tau_r}{\tau_m - \tau_r} \left( e^{-\frac{t}{\tau_m}} - e^{-\frac{t}{\tau_r}}  \right)$, with time constants $\tau_m$ and $\tau_r$.
Here $\ast$ denotes convolution and $\Theta(\cdot)$ is the Heaviside step function, i.e.\ $\Theta(x)=1$ for $x\ge 0$ and $0$ otherwise.

Spike trains were generated using the following method. We used an exponential dependence between the membrane potential and firing rate, such that the instantaneous rate of neuron $k$ at time $t$ is given by $f_k(t) \;=\; \exp(u_k)$. Spike events were drawn from a Poisson process with rate $f_k(t)$. After each spike, neurons were refractory for a fixed time window of length $t_\text{ref}$.

The bias potential $\vartheta_k(t)$ in Eq.~\eqref{eq:membrane-potential} implements a slow rate adaptation mechanism which was updated according to
\begin{equation}
\tau_{\vartheta}\, \frac{\d \vartheta_k(t)}{\d t} \;=\; \nu_0 - z_k(t)  \;,
\label{eqn:adaptation-current}
\end{equation}
where $\tau_{\vartheta}$ is the time constant of the adaptation mechanism and $\nu_0$ is the desired output rate of the neuron.
In our simulations, the bias potential  $\vartheta_k(t)$ was initialized at -3 and then followed the dynamics given in Eq.~\eqref{eqn:adaptation-current} (see~\cite{kap17} for details).

\subsection{Reward-based synaptic sampling}
\label{sec:reward-based-sampling}

In a reward-based learning framework we assume that the network is exposed to a real-valued scalar
function $r(t)$ that denotes the reward at any moment in time in response to the network behavior. The value function $\mathcal{V}(\bth)$ determines the expectation of $r(t)$ over all possible network states while discounting future rewards, i.e.
$\mathcal{V}(\bth) \;=\; \expect{ \int_{0}^\infty e^{-\frac{\tau}{\tau_e}} \,\rtau \; \d \tau }$, with discounting time constant $\tau_{e}$ and $\expect{\cdot}$ denotes the expectation over all possible network responses.

The gradient $\ddthetai \log \mathcal{V}(\bth)$ can be estimated for the network model outlined above using standard reward-modulated learning rules with an eligibility trace (see~\cite{kap17} for details)
\begin{equation}
  \frac{\d e_{i}(t)}{\d t} \;=\; - \frac{1}{\tau_{e}} e_{i}(t) \;+\; w_{i}(t) \, \hz_{\prei}(t) \,  (\zkt -f_{\posti}(t))\;, \label{eqn:eligibility-trace}
\end{equation}
where $\tau_e$ is the time constant of the eligibility trace.
Recall that $\preidef$ denotes the index of the presynaptic neuron and $\postidef$ the index of the postsynaptic neuron for synapse $i$.
In Eq.~\eqref{eqn:eligibility-trace} $\zkt$ denotes the postsynaptic spike train, $\fkt$ denotes the instantaneous firing rate of the postsynaptic neuron and $w_{i}(t) \, \hz_{\prei}(t)$ denotes the postsynaptic potential under synapse~$i$.

This eligibility trace Eq.~\eqref{eqn:eligibility-trace} is multiplied by the reward $r(t)$ and integrated in each synapse $i$ using a second dynamic variable
\begin{equation}
\frac{\d g_{i}(t)}{\d t} \;=\; - \frac{1}{\tau_{g}} g_{i}(t) \;+\;\left( \frac{\rt}{\rhatt} + \alpha \right) \, e_{i}(t) \;,
\label{eqn:gradient-est}
\end{equation}
where $\rhatt$ is a low-pass filtered version of $\rt$ with time constant $\tau_g$.
The variable $g_{i}(t)$ combines the eligibility trace $e_{i}(t)$ and the reward $\rt$ in a temporal average.
$\alpha$ is a constant offset on the reward signal.
This parameter can be set to an arbitrary value without changing the stationary dynamics of the model~\cite{kap17}.
In our simulations, this offset $\alpha$ was chosen slightly above $0$ ($\alpha=0.02$) such that small parameter changes were also present without any reward.
The variable $g_{i}(t)$ realizes an online estimator for $\ddthetai \log \mathcal{V}(\bth)$ \cite{kap17}.

\begin{table}
  \centering
  \caption{Parameters of the neuron and synapse model Eqs.~\eqref{eq:membrane-potential}-\eqref{eqn:std-synapse}.}\label{tab:params}
  \begin{tabular}{ccl} \toprule
    \textit{symbol}                   & \textit{value} & \textit{description}                                                                      \\ \midrule
    $\tau_r$                          & 2~ms           & time constant of EPSP kernel (rising edge)                                                \\
    $\tau_m$                          & 20~ms          & time constant of EPSP kernel (falling edge)                                               \\
    $\tau_e$                          & 1~s            & time constant of eligibility trace                                                        \\
    $\tau_\vartheta=\tau_g$    & 50~s           & time constants for Eq.~\eqref{eqn:adaptation-current} and Eq.~\eqref{eqn:gradient-est}  \\
    $\nu_0$                           & 5~Hz           & desired output rate                                                                       \\
    $t_\text{ref}$                    & $5~ms$         & refractory time                                                                           \\
    $T$                               & 0.1            & temperature                                                                               \\
    $\alpha$                          & 0.02           & offset to reward signals                                                                  \\
    $\beta$                           & $10^{-5}$      & learning rate                                                                             \\
    $\mu$                             & 0              & mean of prior                                                                             \\
    $\sigma$                          & 2              & std of prior                                                                              \\ \bottomrule
  \end{tabular}
\end{table}

Putting it all together, by plugging Eq.~\eqref{eqn:gradient-est} into Eq.~\eqref{eq:sde-reduced} the synaptic parameter changes at time $t$ are given by
\begin{equation}
  \begin{split}
    d \thi \;=\; & \beta \, \left( \frac{1}{\sigma^2}\left( \mu - \thi \right) \, + \, g_{i}(t)  \right)  dt  \;+ \\
             & \sqrt{2 \beta T} \, d \wiener_{i}(t) \;.
  \end{split}
  \label{eqn:std-synapse}
\end{equation}

Eqs.~\eqref{eq:thetamap} and~\eqref{eq:membrane-potential}-\eqref{eqn:std-synapse} conclude the neuron and synapse dynamics used in our simulations.
The parameter values are given in Table~\ref{tab:params}.

\subsection{Random Reallocation of Synapse Memory}
\label{sec:ran_rew}

In the original synaptic sampling model, outlined above, whenever a synapse $i$ is disconnected (when $\theta_i \leq 0$), it undergoes a random walk according to Eq.~\eqref{eq:sde-reduced} until $\theta_i$ again becomes larger than zero and the synapse reappears. The dynamics of synapses that are disconnected also become independent of the network activity and are therefore not influenced by the pre- and post-synaptic spike trains, since the eligibility trace Eq.~\eqref{eqn:eligibility-trace} vanishes. Nevertheless, synapses need to be updated even when they are not used which wastes memory and CPU time. In a typical simulation of synaptic sampling, where the majority of synapses are non-functional most of the time, this overhead may even dominate the simulation. Here, we discuss a more efficient approach for synaptic rewiring called \emph{random reallocation of synapse memory}.

It has been previously noted that the synaptic sampling dynamics can be replaced by a more efficient approach for online rewiring of neural networks  \cite{bel17}. The theoretical analysis there has shown that the original synaptic sampling formulation, with convergence to a stationary distribution $p^*(\bth)$, can be combined with a hard constraint on the network connectivity such that at any moment in time a fixed number of connections $M$ is functional, i.e. $|\bth > 0| = M$. In this modified version of network rewiring, whenever a connection becomes non-functional another synapse is randomly reintroduced to keep the total number of synapses constant. Thus, non-functional synapses do not need to be simulated and therefore don't waste memory or CPU time. It has been shown that this more efficient rewiring approach also leads to a stationary distribution of network configurations, that is identical to the original posterior $p^*(\bth)$ confined to the manifold of the parameter space that fulfills the constraint $|\bth > 0| = M$ (see \cite{bel17} for details). This rewiring strategy has already been successfully applied to deep learning~\cite{bel17} and implemented on the SpiNNaker~2 prototype chip~\cite{liu2018memory}.

Here, we used a similar rewiring approach to the one in \cite{bel17}. However, an additional limitation on the rewiring scheme comes from the memory model of the software framework. In our implementation, each neuron maintains a table of its post-synaptic targets (see Section \ref{sec:mem_mod} for details). Therefore, the free space of synapses that become disconnected can most efficiently be reassigned to another postsynaptic target of the same presynaptic neuron. Consequently, we decided to use a connectivity constraint that assures that the \emph{fanout of each neuron} is constant throughout the simulation. This is simply achieved by immediately reconnecting each synapse that becomes non-functional to a new randomly chosen postsynaptic target. Since drawing random numbers becomes efficient due to the random number generator (Section~\ref{sec:ran_num}), this approach has little computational overhead.

Our results from the prototype chip presented in Section~\ref{sec:implementation_results} suggest, that random reallocation increases the effective usage of the hardware, the number of active synapses in the network, and also accelerates the exploration of the parameter space, leading to faster convergence to the stationary distribution.
Interestingly, the connectivity constraint used here is somewhat similar to analog neuromorphic systems which contain synaptic matrices fixedly assigned to postsynaptic neurons with only the presynaptic sources flexible to some degree \cite{noack2010biology}. Rewiring in such a setup has to operate `postsynaptic-centric' and similar to our approach has a fixed number of synapses per postsynaptic neuron \cite{george2015event}.

\section{Implementation of Synaptic Sampling on the SpiNNaker 2 Prototype}
\label{sec:software}

The software implementation of this model is optimized regarding computation time, memory, power consumption and scalability, in order to bridge the gap between state-of-the-art biologically plausible neural models and efficient execution of the model in hardware. This is explained in more detail in the following.  

\subsection{Numerical Optimizations}\label{sec:num_opt}

\paragraph{Reducing computation time with hardware generated uniform random numbers}
The synaptic sampling model draws one random number for each synapse in each simulation time step (1 ms).
Since thousands of synapses are simulated in each core, random number generation could dominate the computation time.
As described in Section~\ref{sec:synapse_model}, the Wiener process requires Gaussian random numbers to be generated.
But as described in Section~\ref{sec:ran_num}, only uniform random number can be generated by the accelerator.
As shown in Table~\ref{tab:clock_cycles}, the generation of a pseudo Gaussian random number with Box-Muller transform~\cite{box58} in software requires 172 clock cycles.
One option could be to convert the hardware generated uniform random number into Gaussian random number with Inverse CDF method~\cite{hor03} and look-up table, which reduces the computation time to 21 clock cycles.
However, analytical and numerical studies have found that for the simulation of Wiener process, Gaussian random numbers can be replaced by uniform random numbers without affecting model performance~\cite{dun91}.
The generation of a uniform random number in software with Marsaglia RNG~\cite{mar03,hop14} requires 42 clock cycles,
whereas with hardware it takes only 5 clock cycles,
including fetching the integer random number from the accelerator and converting it to floating point type in the range of 0 to 1.

  \begin{table}[htb]
    \begin{center}
      \renewcommand{\arraystretch}{1.1}
      \caption{Computation time for random number generation and exponential function}\label{tab:clock_cycles2}\label{tab:clock_cycles}
      
      Computation time for random number generation
      \begin{footnotesize}
        \begin{tabular}{p{5.2cm}p{1.9cm}}\toprule
          Random number type                          & \#clock cycles   \\ \midrule
          Gaussian (software, Box-Muller Transform)   & 172              \\
          Gaussian (hardware, Inverse CDF, optimized) & 21               \\
          Uniform (software, Marsaglia)               & 42               \\
          Uniform  (Hardware)                         & 5                \\ \bottomrule \\
        \end{tabular}
        
        \end{footnotesize}
              Computation time for exponential function
              \begin{footnotesize}
              \begin{tabular}{p{5.2cm}p{1.9cm}} \toprule
                Exponential function                         & \#clock cycles \\ \midrule
                Software (floating point, Newlib)            & 163            \\
                Software (fixed point, hardware emulation)                & 104            \\
                Hardware (fixed point, precision not enough) & 6              \\
                Hardware (conversion from and to float)      & 15             \\ \bottomrule
              \end{tabular}
      \end{footnotesize}
    \end{center}
  \end{table}

\par
\paragraph{Reducing computation time with exponential function accelerator}
In the synapse model, the parameter $\theta$ of each synapse accumulates small changes in each time step.
The exponential function accelerator, which calculates the exponential function within 6 clock cycles (Section~\ref{sec:exp_fun}), uses a fixed-point data type whose precision is not enough for this model, because the change of $\theta$ would be rounded to zero.
Calculating a floating point exponential function with software libraries like Newlib takes 163 clock cycles.
Since high precision is only necessary for storing the small change of $\theta$, but not necessary for calculating intermediate variables like $w$, $\theta$ can be stored as floating point in memory, and when calculating $w$ with exponential function, $\theta$ can be converted to fixed point and calculated with the exponential function accelerator.
The result is then converted back to floating point.
Simulations show that the performance of the model is not affected.
This reduces the computation time to 15 cycles with 6 cycles required by the hardware accelerator and 9 additional cycles for the conversion of data type.
For the sake of comparison, emulation of exponential accelerator in software takes 95 cycles instead of 6~\cite{par17}.
Thus, with conversion of data type, this approach would take 104 cycles with software (Table~\ref{tab:clock_cycles2}).

\paragraph{Reducing memory footprint with 16-bit floating point data type}
In order to simulate more synapses with limited memory, which is the case when the synapse parameters are stored in SRAM (see Section~\ref{sec:loc_com}), the single precision floating point with 32 bits can be converted into half precision floating point with 16 bits.
For each synapse $i$, three parameters need to be stored in memory: \emph{eligibility trace} $e_i$, \emph{estimated gradient} $g_i$ and \emph{synaptic parameter} $\theta_i$.
Simulations show that converting $e_i$ and $g_i$ to half precision does not affect the model performance.

\subsection{Local Computation}\label{sec:loc_com}
By avoiding external DRAM access and instead storing all parameters and state variables of the model locally in SRAM, both energy and computation time can be saved.
\begin{figure}[htb]
  \centering
  \includegraphics[width=0.48\textwidth]{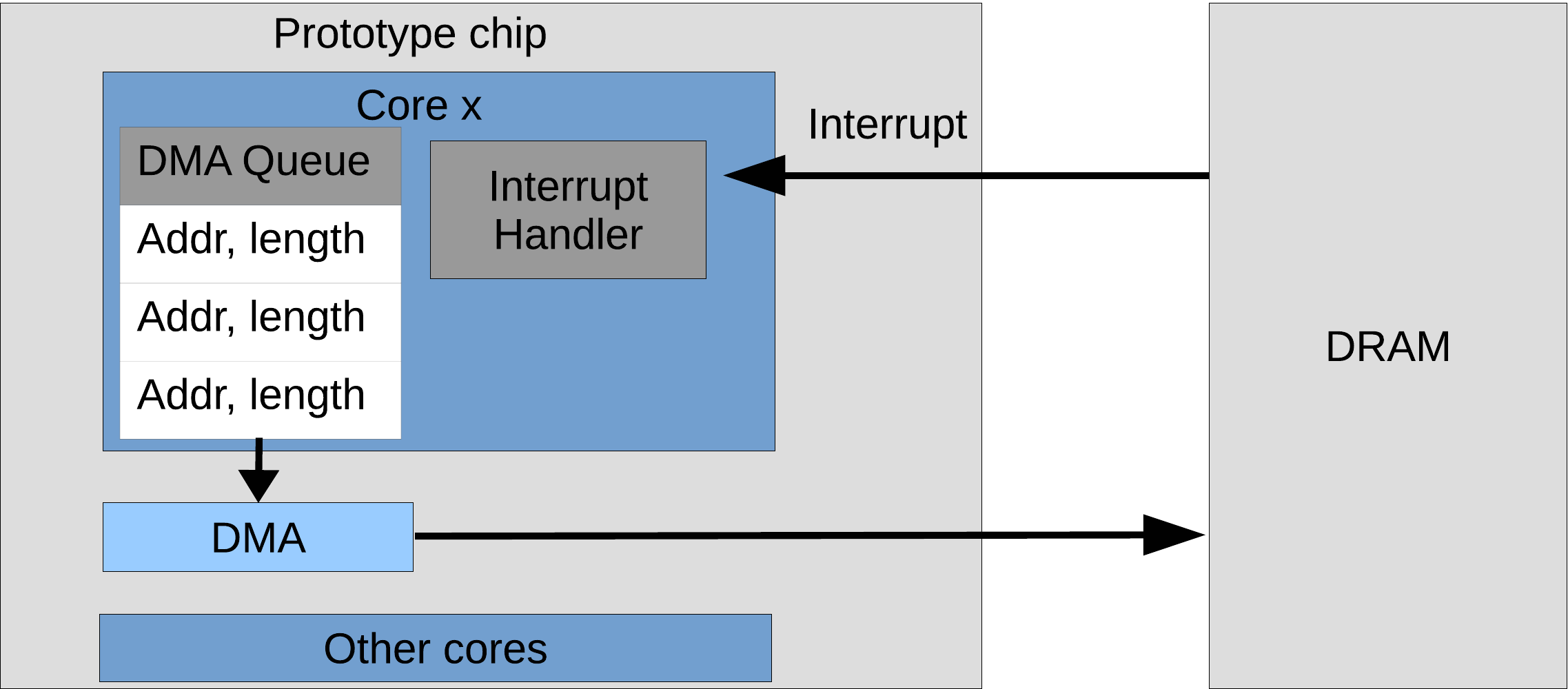}
  \caption{\label{fig:local_computation} The time and energy consuming interaction between the prototype chip and the DRAM chip, which can be saved by storing data locally in SRAM.}
\end{figure}

To read (write) data from (to) the off-chip DRAM, the core sends a read (write) request which is first stored in a DMA (Direct Memory Access) queue in software, then sent to the DMA unit, and at last sent to the DRAM\@.
When the read (write) process is complete, an interrupt is triggered and an interrupt handler is called, which, in case of read request, processes the data from DRAM\@.
Then the next read/write request in the queue is sent to DMA (Fig.~\ref{fig:local_computation}).
Since the DRAM access is time consuming, the software can let DMA run in background and continue with other tasks.
When the read/write process is complete, the core stops with the current task, handles the interrupt and then resumes the stopped task after the interrupt handler is complete.
Although this saves computation time compared to waiting for the read/write process to complete, it still has the following drawbacks:
\begin{enumerate}
  \item Retrieving all synapse parameters in each time step, which is necessary in this model, could easily saturate DRAM bandwidth especially in the scaled up case with tens of cores per chip~\cite{mantas18,pai13}.
  \item The energy consumption of DRAM access can be two orders of magnitudes higher than SRAM access~\cite{Han15}.
  \item This only works if the other tasks are independent from the data being fetched.
  \item Managing the DMA queue and calling the interrupt handler still consumes computation time, which becomes a problem when memory is frequently accessed.
\end{enumerate}

The drawback when not using external DRAM is the limited memory space available in SRAM\@.
This is not a problem for this model, since on the one hand the required memory is reduced with 16-bit floating point (Section~\ref{sec:num_opt}), and on the other hand due to the complexity of the model, the number of synapses per core is limited by computation as is shown in Section~\ref{sec:net_top}.

\subsection{Memory Model}\label{sec:mem_mod}

\begin{figure}[htb]
\centering
 \includegraphics[width=0.5\textwidth]{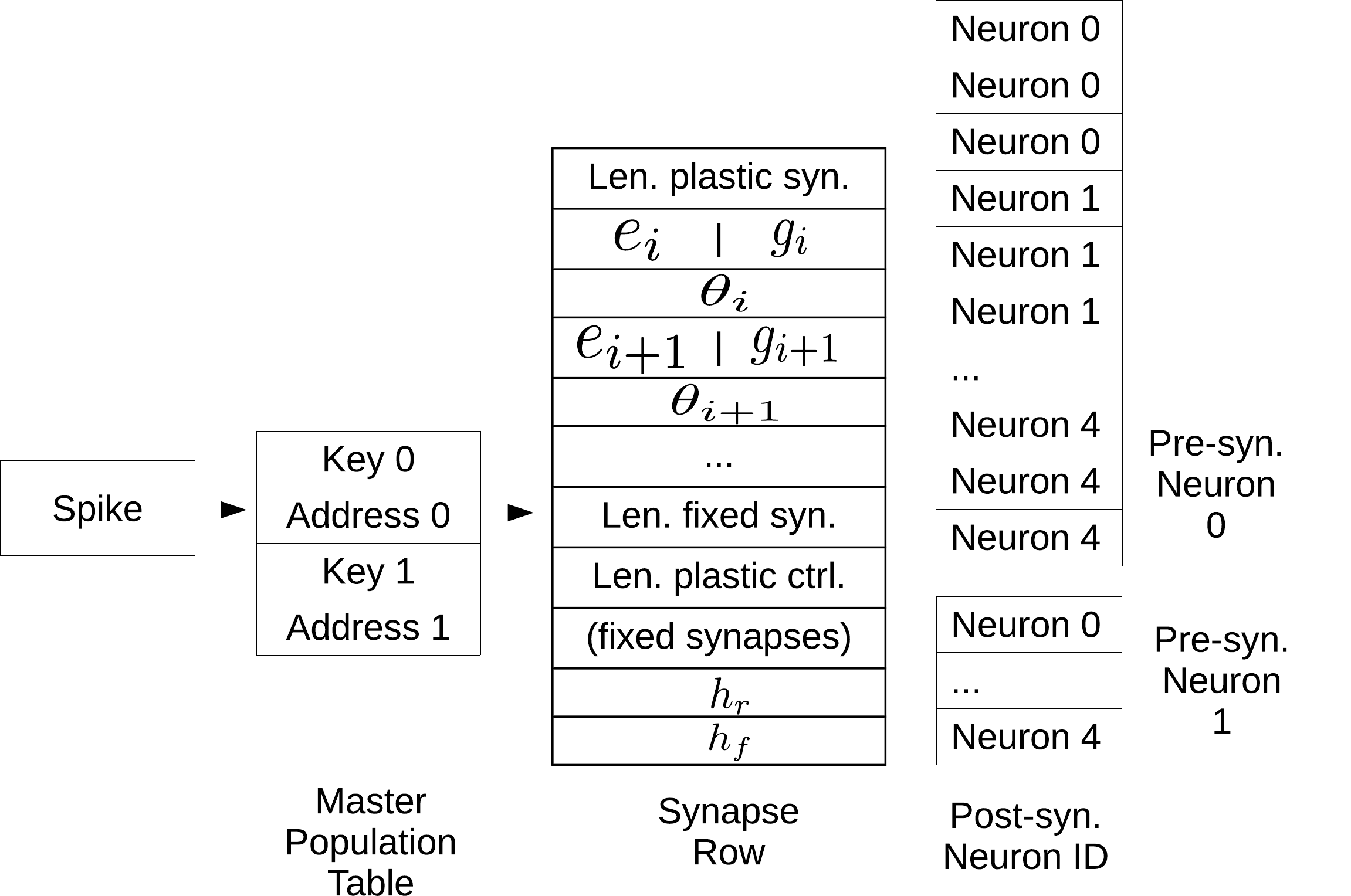}
\caption{\label{fig:mem_mod} Memory model with master population table, synapse rows and postsynaptic neuron ID.}
\end{figure}

The memory model (Fig.~\ref{fig:mem_mod}) of this work is based on the software for the first generation SpiNNaker system~\cite{rhodes2018spynnaker}.
The spike packet contains the ID of the presynaptic neuron.
The master population table contains keys which are presynaptic neuron IDs.
Each key is 4 bytes long and is stored together with the 4 byte starting address of the synapse parameters for the presynaptic neuron.
These synapse parameters are stored in a contiguous memory block called synapse row.
Each row is composed of 4-byte words.
For each presynaptic neuron, the first word is the length of the plastic synapse region.
In our implementation, the plastic synapse region consists of 8-byte blocks with 2 bytes for $e_i$, 2 bytes for $g_i$ and 4 bytes for $\theta_i$.
After the plastic synapse region there is one word for the length of fixed synapse region.
The next word is the length of the plastic control region which stores special parameters needed by the plasticity rules.
In this work this region is used to store the parameters for the PSP kernel of input spike, e.g.\ $h_{r}$ and $h_{f}$ (corresponding to the time constants $\tau_m$ and $\tau_f$).
Since the PSP kernel of the incoming spike is the same for all synapses of the same presynaptic neuron, the parameters for the PSP kernel are shared in order to reduce memory footprint.
After the word for the length of plastic control region follow the parameters for fixed synapses.

The synapse parameters should also include the index of the postsynaptic neuron.
One way to implement this is to add a 4-byte word for each postsynaptic neuron in addition to the 8 bytes for $e_i$, $g_i$ and $\theta_i$, which is the case in the original SpiNNaker software framework.
Alternatively, since in this network all input neurons have the same fanout, the indexes are stored in a 2-d array (Post-syn. Neuron ID in Fig. \ref{fig:mem_mod}), where the column index stands for the presynaptic neuron ID and the entries represent the postsynaptic neuron IDs. Each entry represents a synapse and occupies one byte, supporting maximum 256 target neurons per core.
Since multiple synapses are allowed between a pair of neurons, the ID of a postsynaptic neuron can appear multiple times in each column of the 2-d array.
In general, depending on application, one of the two approaches can be chosen.

The master population table, synapse rows and postsynaptic neuron ID are arrays generated by each core after the network configuration is specified.
Each core generates its own data in a distributed way instead of having a centralized host PC generating data for all cores.
This, combined with local computation (Section~\ref{sec:loc_com}), drastically reduces the time for data generation and transmission of data from host PC to chip, which could make up significant amount of total simulation time especially in the case of large systems~\cite{sha13,vanAlbada2018performance}.

\subsection{Program Flow and SpiNNaker Software Framework Integration}\label{sec:pro_flow}

\begin{figure}[htb]
  \centering
  \includegraphics[width=0.48\textwidth]{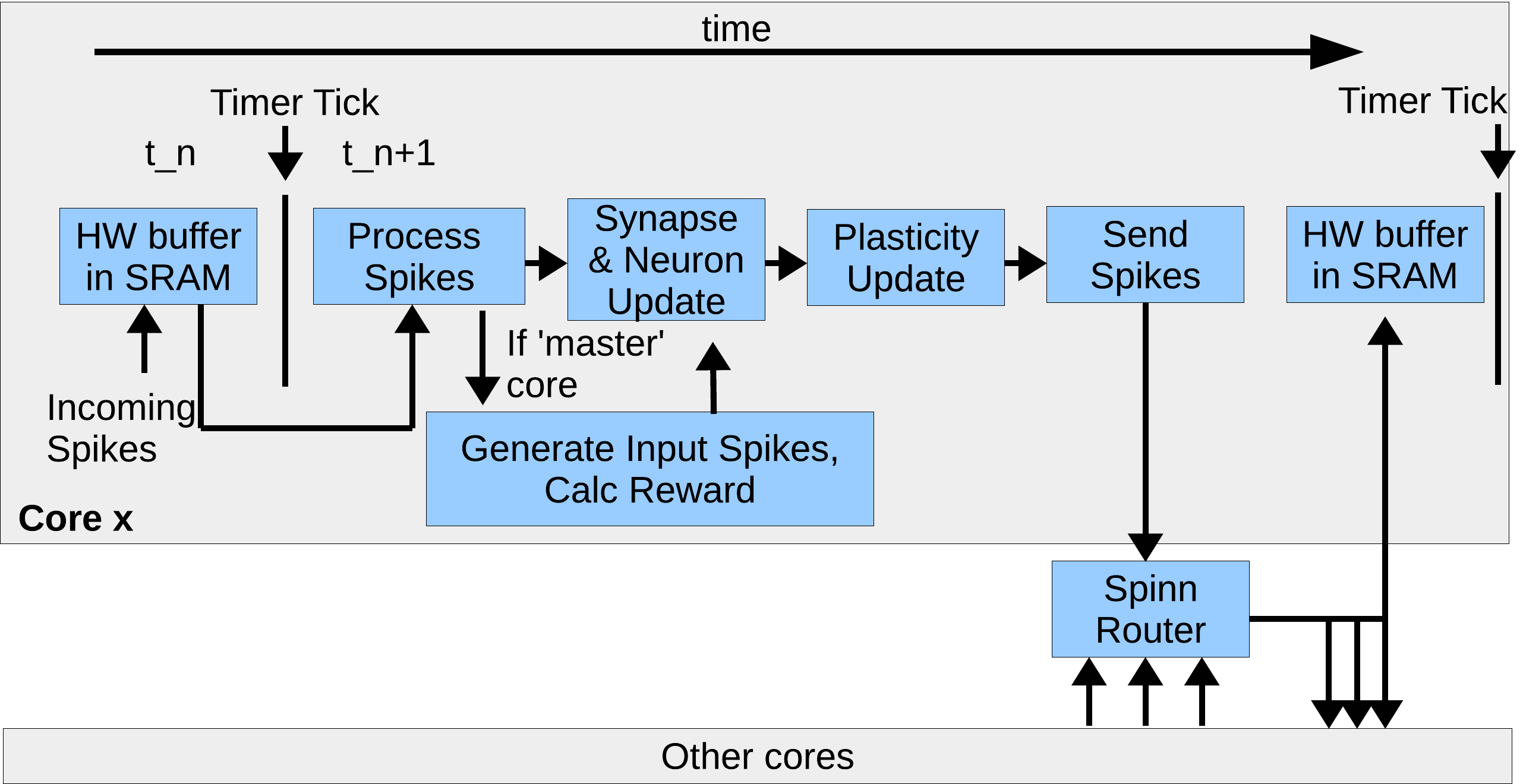}
  \caption{%
    SpiNNaker software framework. Each simulation time step $t_n$ is triggered by the timer tick interrupt. At the end of the time step, the spikes are sent to the SpiNNaker router which then multicasts the spikes to other cores.
  }\label{fig_spinn_software}
\end{figure}

The SpiNNaker system employs parallel computation to run large scale neural simulations in real time.
Although the prototype chip consists of only 4 cores, the software implementation of the synaptic sampling model is integrated into the SpiNNaker software framework allowing for scaling up onto larger systems.
The design of the program flow is based on~\cite{rhodes2018spynnaker}.

The timer tick signal of the ARM core is used to trigger each time step in real time.
The length of a time step can be arbitrarily chosen.
For this implementation, one time step is one millisecond.
The timer tick signal triggers an interrupt.
Then the handler of the interrupt is called and processes the incoming spikes from the last time step, which are stored in a hardware buffer in SRAM\@.
In this step, for each incoming spike, first the starting memory address of its corresponding synapse parameters is found in the master population table, then the synaptic weights of the activated synapses in the synapse row are added to the synaptic input buffers of the target neurons.

For the network model implemented in this work (Section~\ref{sec:net_top}), one of the cores, the ``master core'', then simulates the environment that computes the global reward signal.
All cores continue with the synapse update and neuron update, which integrate the synaptic weight onto the membrane potential of the postsynaptic neuron.
Next, the synaptic plasticity update is performed, as now all required information is available, i.e.\  incoming spikes, neuron states and global reward. 

At last, the spikes of the neurons in each core are sent to the SpiNNaker router, which then multicasts the spikes to the cores containing the corresponding postsynaptic neurons.
The SpiNNaker router~\cite{nav15} allows for fast multicast of small packets, which is key to efficient spike communication for many-core neuromorphic systems like SpiNNaker.
The distributed computation, synchronization with timer tick and communication with the SpiNNaker router allows for scaling up the model implementation onto large systems consisting of millions of cores.


\section{Results}
\label{sec:results}
In the following we show how the hardware accelerators and numerical optimizations reduce the computation time for one plasticity update of the synaptic sampling model.
Then, we implement a network model that performs reward-based synaptic sampling on the SpiNNaker 2 prototype, for which we also provide power and energy measurements.

\subsection{Computation Time of Plasticity Update}\label{sec:comp_time}

\begin{center}
  \begin{footnotesize}
    \begin{table}[htb]
      \renewcommand{\arraystretch}{1.1}
      \caption{Number of clock cycles for plasticity update}\label{tab:syn_pro}
      \begin{tabular}{p{3.2cm}p{1.9cm}p{1.9cm}}  \toprule
                                 & HW Accelerator & only Software \\ \midrule
        Random number generation & 5              & 42            \\
        Exponential function     & 15             & 104           \\
        Rest                     & 90             & 90            \\
        Total                    & 110            & 236           \\
        (RNG + EXP) / Total      & 18\%  & 62\% \\ \bottomrule
      \end{tabular}
    \end{table}
  \end{footnotesize}
\end{center}

As shown in Section~\ref{sec:num_opt} the generation of a uniform distributed random number takes 5 clock cycles with hardware accelerator and 42 clock cycles with software.
The floating point exponential function with exponential accelerator and conversion of data type takes 15 clock cycles, whereas the same algorithm in software takes 104 clock cycles.
The rest of the plasticity update of a synapse takes 90 clock cycles.
In total, the plasticity update takes 110 clock cycles with hardware accelerators and the equivalent implementation with only software takes 236 clock cycles (Table~\ref{tab:syn_pro}).
For this application, the hardware accelerators result in a speedup of 2 regarding the number of clock cycles.
Considering the increase of clock frequency from 200 MHz in SpiNNaker~1 to 500 MHz in the current prototype chip, in total a speedup factor of~5 is achieved.
In the plasticity update, the computation time for random number generation and exponential function reduced  from  62\% to 18\%.

\subsection{Network Description}\label{sec:net_top}
\begin{figure}[htb]
  \centering
  \includegraphics[width=0.48\textwidth]{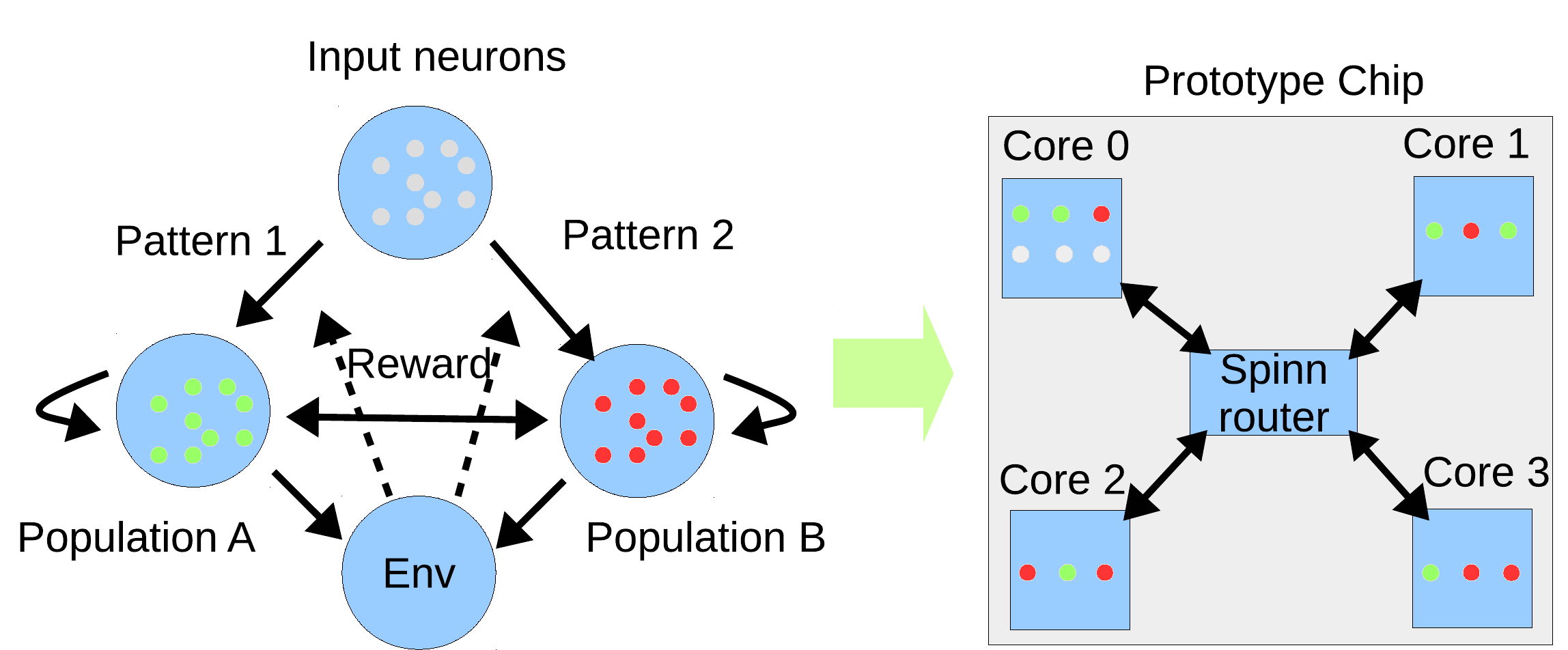}
  \caption{%
  	Illustration of the network topology (left) and its mapping to the prototype chip (right).
  }\label{fig:topology} 
\end{figure}


Fig.~\ref{fig:topology} illustrates the network topology and the mapping to the prototype chip. The network consists of 200 input neurons which are all-to-all connected to 20 neurons with plastic synapses. Multiple synapses between each pair of neurons are allowed. In this implementation 3 synapses between each pair of neurons are initiated, resulting in 200 x 20 x 3 = 12000 plastic synapses. 2 spike patterns are encoded in the spike rate of the input neurons and are sent to the hidden neurons (see Fig. \ref{fig:spikes}).
The 20 hidden neurons are divided into two populations (A and B). The output spikes of the hidden neurons are sent to the environment (Env), which evaluates the global reward. A high reward is obtained if input pattern 1(2) is present and the mean firing rate of population A(B) is higher than population B(A). The global reward is sent back to the network and shapes the plastic synapses between the input neurons and the two populations. The goal is to let the two populations `know' which spike pattern they represent and signal this with a high firing rate when their pattern is present. In addition to the feedforward input, hidden neurons receive lateral inhibitory synapses that are initiated to fixed random weights between each pair of hidden neurons.

The network is mapped to the prototype chip with each core simulating 5 neurons from the two populations (see Fig.~\ref{fig:topology}).
The first core ("master core") also generates the input spikes and evaluates the reward. The 200 input neurons lead to $200 \times 5 = 1000$ pairs of neurons in each core.

The profiling results in section \ref{sec:comp_time} provide the computational aspect when assigning the number of synapses to simulate on each core. The ARM Cortex M4F core used in this prototype chip is configured to run at 500 MHz, which means 500 000 clock cycles are available in each time step (1 ms). 
The computation for one time step without plasticity update takes ca. 45 000 clock cycles for core 0 and 40 000 clock cycles for the other cores. 
Since each plasticity update takes 110 cycles with hardware accelerators and 236 cycles without hardware accelerators, the theoretical upper limit for the number of synapses per core is ca. 4 100 with hardware accelerators and ca. 1 900 without hardware accelerators.

In terms of memory, the prototype chip has 64 kB Data Tightly Coupled Memory (DTCM) per core, for all initialized data, uninitialized data, heap and stack. By checking the binary file size after compilation, the maximum number of synapses is estimated as 4 700. Thus, this model is limited by computation rather than memory (see table \ref{tab:net_pro}).

\begin{center}
  \begin{footnotesize}
    \begin{table}[htb]
      \renewcommand{\arraystretch}{1.1}
      \caption{Maximum Number of Synapses per Core}\label{tab:net_pro}
      \begin{tabular}{p{2.5cm}p{2.5cm}p{2.5cm}}  \toprule
                     & Core Memory Constraint & Real Time Constraint \\ \midrule
        With Accelerators          & 4 700              & 4 100        \\
        Without Accelerators       & 4 700             &  1 900        \\ \bottomrule
        
      \end{tabular}
    \end{table}
  \end{footnotesize}
\end{center}

In the implementation, 3 000 plastic synapses per core are simulated, in order to ensure the stability of the software. Since 3 000 plastic synapses can be simulated in each core, each pair of neurons has 3 plastic synapses.
Note that this is only the initial configuration.
Due to random reallocation of synapse memory, the postsynaptic neuron could change, so that not each single pair of neurons has 3 plastic synapses.


\subsection{Implementation Results}
\label{sec:implementation_results}

The usability of the network is demonstrated in a closed-loop reinforcement learning task implemented with 4 ARM cores. The generation of input spikes and evaluation of output spikes are also implemented on chip.

\begin{figure}[htb]
  \centering
  \includegraphics[width=0.5\textwidth]{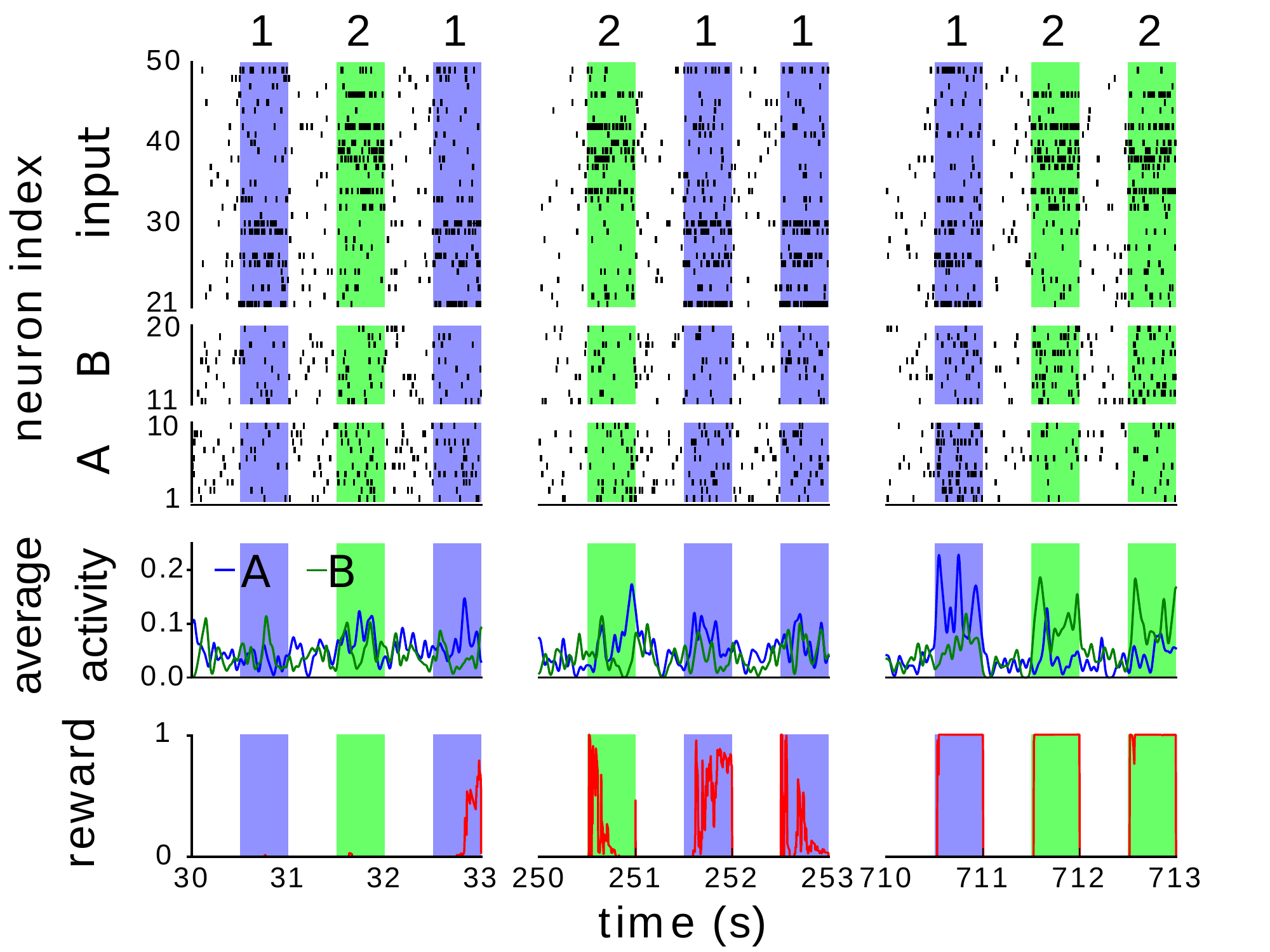}
  \caption{%
  	Network activity and reward throughout learning. Shaded areas indicate the presented patterns. 	Spike trains (top) of the two populations and input spikes. 30 neurons were randomly chosen from the 200 inputs.
  }\label{fig:spikes}
\end{figure}

As shown in Fig.~\ref{fig:spikes}, the 200 input neurons send two spike patterns in random order. Each spike pattern lasts for 500 ms. Resting periods of 500 ms are inserted between two pattern presentations, where the input neurons only send random spikes with low firing rate representing background noise. The numbers at the top of Fig.~\ref{fig:spikes} and shaded colored areas indicate which pattern is present. As discussed above, the 20 neurons are divided into 2 populations (A and B), each representing one of the two patterns. Neuron~1 to neuron~10 belong to population A, neuron~11 to neuron~20 belong to population B. In the second row of Fig.~\ref{fig:spikes}, blue and green curves represent population firing rates of A and B, respectively. The firing rates were obtained with a Gaussian filter ($\sigma = 20~\text{ms}$) applied to the raw spike trains. The goal of learning is to let population A fire at a higher rate when pattern 1 is present and let population B fire at a higher rate when pattern 2 is present.

\begin{figure}[htb]
  \centering
  \includegraphics[width=0.5\textwidth]{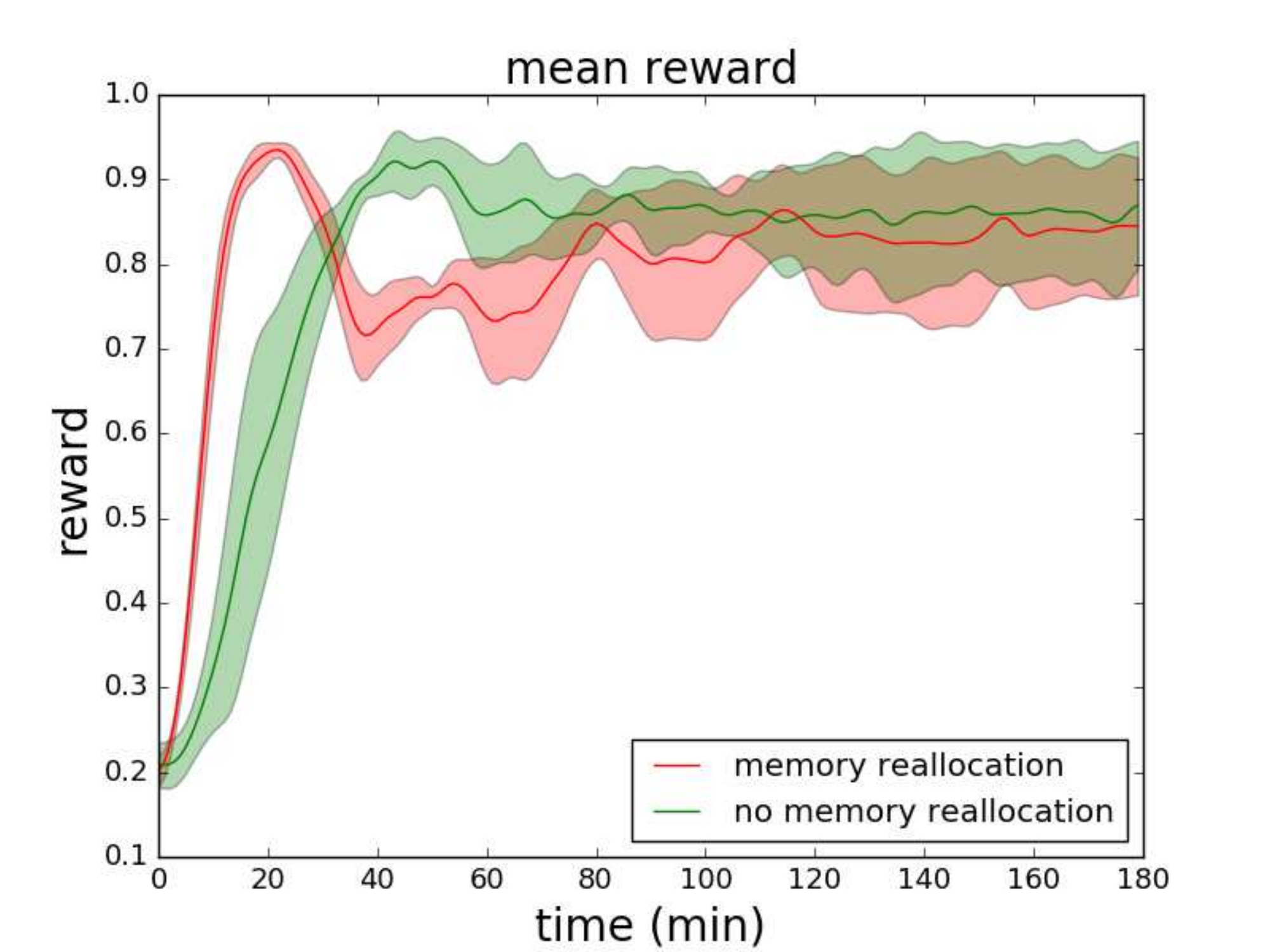}
  \caption{%
    Time-averaged reward over throughout learning for networks with (red) and without (green) random reallocation of synapse memory.
  }\label{fig:reward}
\end{figure}

Fig.~\ref{fig:reward} shows the evolution of the mean reward with and without random reallocation of synapse memory (see Section \ref{sec:ran_rew}). The mean reward in each minute is low-pass filtered with a Gaussian kernel with $\sigma = 2~\text{min}$. Averages over 5 independent trial runs using the true random number generator are shown with solid lines, shaded areas indicate standard deviations. The reward is normalized to the theoretically maximum reachable reward.
At learning onset the two populations respond randomly to input spike patterns and the reward is low. The synaptic weights explore the parameter space with the random process guided by the global reward as described in Section~\ref{sec:synapse-model}. Over time, the network learns the desired input/output mapping and the reward increases.  After ca.~10 minutes of training, the two populations learn to respond correctly to the two spike patterns with the firing rate of one population higher than the other when the corresponding spike pattern is present, and reward becomes high. Our results show that the reward increases much faster with reallocation due to the accelerated exploration of the parameter space. After the reward reaches a high value, the network continues exploration and the reward might fluctuate while the network searches for equally good network configurations.

\subsection{Power and Energy Measurement Results}\label{sec:power_energy}

\begin{center}
  \begin{footnotesize}
    \begin{table}[htb]
      \renewcommand{\arraystretch}{1.1}
      \caption{Power and Energy Consumption}\label{tab:pow_ener}
      \begin{tabular}{p{1.4cm}p{1.9cm}p{1.9cm}p{1.9cm}}  \toprule
                     & with DRAM, no Accelerator & no DRAM, no Accelerator & no DRAM, with Accelerator \\ \midrule
        Power  (mW)        & 285              & 225      &  225    \\
        Time   (ms)       & 1.58             & 1.58     &  0.76    \\
        Energy (\(\mu\)J)       & 450.3             & 355.5      &  171    \\ 
        Reduction of Energy        & 0\%                  & 21\%       &   62\% \\ \bottomrule
        
      \end{tabular}
    \end{table}
  \end{footnotesize}
\end{center}

The optimizations described in section \ref{sec:software} result in considerable reduction of power and energy consumption. To show the benefit of the optimizations, power and energy consumption is measured in three cases. First, the synapse rows are stored in the external DRAM memory, and the exponential function and random number generation are done only with the software running on ARM core. Second, the synapse rows are stored in the local SRAM memory, and the exponential function and random number generation are still only done with the software running on ARM core. At last, the synapse rows are stored in the local SRAM memory, and the exponential function and random number generation are done with the hardware accelerators. For this measurement, the software is run without random reallocation of synapse memory. As summarized in table \ref{tab:pow_ener}, the power and energy consumption is reduced by local computation without external DRAM and reduction of computation time. 

First, the memory footprint is optimized by employing 16-bit floating point data type and the compact arrangement of memory model described in sections \ref{sec:num_opt} and \ref{sec:mem_mod}. The random reallocation described in section \ref{sec:ran_rew} increases the effective number of synapses which is otherwise only achievable with external memory like DRAM. The reduction of memory footprint allows for local computation with SRAM, as described in section \ref{sec:loc_com}. Switching off DRAM allows for a reduction of power consumption by 21\%, from 285 mW to 225 mW.

In addition, as summarized in section \ref{sec:comp_time}, the computation time for each plasticity update is reduced by 53.4\%. Without the hardware accelerators, simulating the network with 3 000 plastic synapses per core for one time step (1 ms) takes 1.58 ms, losing the real time capability. With the hardware accelerators, the simulation of one time step is finished within 0.76 ms. To measure the energy consumption, the length of the time step is chosen to be the minimum required for each time step to finish, i.e. 1.58 ms for without accelerators and 0.76 ms for with accelerators. The reduction of computation time for plasticity update reduces the energy consumption for one time step by 51.9\%, from 355.5 \(\mu\)J to171 \(\mu\)J .

In total, the energy consumption for the simulation of the network for one time step is reduced by 62\%, from 450.3 \(\mu\)J to 171 \(\mu\)J, making the system attractive for mobile and embedded applications.


\section{Discussion}
\label{sec:discussion}
In the following we discuss how the implementation of the reward-based synaptic sampling model would scale for larger networks on the final SpiNNaker~2 system.
Finally, we argue about the possiblility to realize this network model on SpiNNaker~1 and other neuromorphic platforms with learning capabilities.

\subsection{Scalability}
The SpiNNaker architecture was designed for the scalable real-time simulation
of spiking neural networks with up to a million cores \cite{fur14}.
SpiNNaker's scalability is based on the multi-cast network for routing of spike
events \cite{nav15} and a software framework for mapping network models onto
the system that has shown to support the simulation of large-scale
neural networks \cite{vanAlbada2018performance}.
Building on this, the reward-based synaptic sampling model can be scaled to future
SpiNNaker~2 systems without major restrictions, i.e.
as our implementation is integrated into the SpiNNaker software framework, the
automatic mapping of larger networks onto many cores and the configuration of
routing tables comes for free.
In principle, with more than 100 cores per chip in SpiNNaker~2
(cf.~Table~\ref{tab:hwspinn}), DRAM bandwidth may become a bottleneck for some
applications, but not in our case, as synapse variables are stored and
processed locally in each core and DRAM is not used. 
Furthermore, a many-chip implementation should not be limited by the communication
bandwith for spike packets between chips, as the reward-based synaptic sampling
model is mainly limited by the computation of the synapse updates and has
rather moderate spike rates (Section~\ref{sec:net_top}).
Still, we remark that, as in any large-scale neuromorphic
hardware system, the fraction of energy consumed for communication will
increase with network size \cite{hasler2013finding} demanding optimized routing
architectures \cite{park2017hierarchical}.

Future work will include simulating larger networks of this type on the full-scale SpiNNaker~2 system with many cores.
Such a scaled-up, real-time version of the synaptic sampling framework, will enable us to explore reward-based learning on high-dimensional input such as dynamic vision sensors \cite{lichtsteiner2008128} or conventional high-density image sensors \cite{henker2007active}. 

\subsection{Comparison with SpiNNaker 1}

Reward-based learning and structural plasticity have been implemented on the SpiNNaker system before~\cite{mantas18}~\cite{bogdan2018structural}. The reward-based synaptic sampling model implemented in this work is more complex because of the need for random number generation and exponential function for each plastic synapse in each time step. In addition, due to the lack of floating point arithmetic, this synapse model would be very hard, if possible at all, to be implemented in the first generation SpiNNaker system, since the change of synaptic weight is very small in each time step and can not be captured by the precision of fixed point format.  

\subsection{Comparison with other neuromorphic platforms}
\label{sec:comparison_platforms}
To the best of our knowledge, there exists today no neuromorphic hardware platform, except SpiNNaker~2, that would be able to directly simulate complex learning rules such as synaptic sampling. Most other approaches have traded off accessible model complexity for a more direct implementation of the neuron dynamics. We discuss here how synaptic sampling could still be emulated on other architectures.
	
Clearly, since synaptic sampling is inherently an online learning model, it cannot be directly implemented on neuromorphic hardware with only static synapses, such as TrueNorth \cite{Merolla668}, NeuroGrid \cite{neurogrid}, HiAER-IFAT \cite{park2017hierarchical}, DYNAPs \cite{dynaps18} and DeepSouth \cite{deepsouth18}. However, the network dynamics could be approximated by alternating short time windows of network simulation and reprogramming synaptic weights by an external device.


Architectures that do support synaptic plasticity on chip, such as Loihi\cite{loihi} and the BrainScales~2 system\cite{PPU}, have so far quite limited weight resolutions (9-bit signed integer on Loihi and 12-bit on BrainScales~2). Since 32-bit fixed-point format was found to be insufficient for this model (cf. section~\ref{sec:num_opt}), it is questionable, even with stochastic rounding, whether synaptic sampling can be implemented with such low weight resolution, and at what cost in performance. Also, in the case of Loihi, the size of the microcode that is allowed for computing synaptic updates is quite limited (e.g. 16 32-bit words). Besides, hardware accelerators for complex functions like the exponential function are not available on these two platforms, which makes the implementation more challenging, especially in the case of Brainscales~2, because the high data rate caused by accelerated operation requires fast execution of learning rules. These restrictions put some doubt on whether complex learning mechanisms, as the one considered here, can be implemented exactly. Also, exact implementation of the synaptic sampling model seems infeasible on neuromorphic hardwares with configurable (but not programmable) plasticity, like ROLLS \cite{rolls15}, ODIN \cite{odin19} and TITAN \cite{titan16} (see \cite{thakur} and \cite{neuromorphic_plasticity_review14} for reviews). However, it might be possible to realize simplified, approximate, versions of synaptic sampling on these neuromorphic platforms.
	


\section{Conclusion}

In this work, a reward-based synaptic sampling model is implemented in the prototype chip of the second generation SpiNNaker system. This real-time online learning system is demonstrated in a closed-loop online reinforcement learning task. 
While hardware features of the future SpiNNaker~2 and its prototypes have already been published, this is the first time learning spiking synapses have been shown on SpiNNaker~2. As shown in sections \ref{sec:intro} and \ref{sec:comparison_platforms}, this is also one of the most complex synaptic learning models ever implemented in neuromorphic hardware. 
The hardware accelerators and the software optimizations allow for efficient neural simulation with regard to computation time, memory and power and energy consumption, while at the same time the SpiNNaker~2 system keeps the full flexibility of being processor based.
For this application, we show slightly more than a factor of 2 speedup of the algorithm compared to a pure software implementation. Coupled with the 2.5~fold increase in clock frequency, we can theoretically simulate 5 times as many synapses of this type in SpiNNaker~2 as in SpiNNaker~1 in the same time span. In addition, we show a reduction of energy consumption by 62\% compared to implementation without the use of hardware accelerators and with external DRAM.

\section*{Acknowledgements}
The research leading to these results has received funding from the European Union Seventh Framework Programme (FP7) under grant agreement no 604102 and the EU's Horizon 2020 research and innovation programme under grant agreements No 720270 and 785907 (Human Brain Project, HBP). In addition, this work was supported by the Center for Advancing Electronics Dresden (cfaed) and the H2020-FETPROACT project Plan4Act (\#732266) [DK]. Furthermore, this work was supported by the Austrian Science Fund (FWF): I 3251-N33.
The authors thank Andrew Rowley, Luis Plana, Alan Stokes and Michael Hopkins for providing the source code of SpiNNaker~1 software. In addition, the authors thank ARM and Synopsis for IP and the Vodafone chair at Technische Universit\"at Dresden for contributions to RTL design.

\bibliographystyle{IEEEtran}
\bibliography{synaptic_sampling_bib}

\begin{IEEEbiography}[{\includegraphics[width=1in,height=1.25in,clip,keepaspectratio]{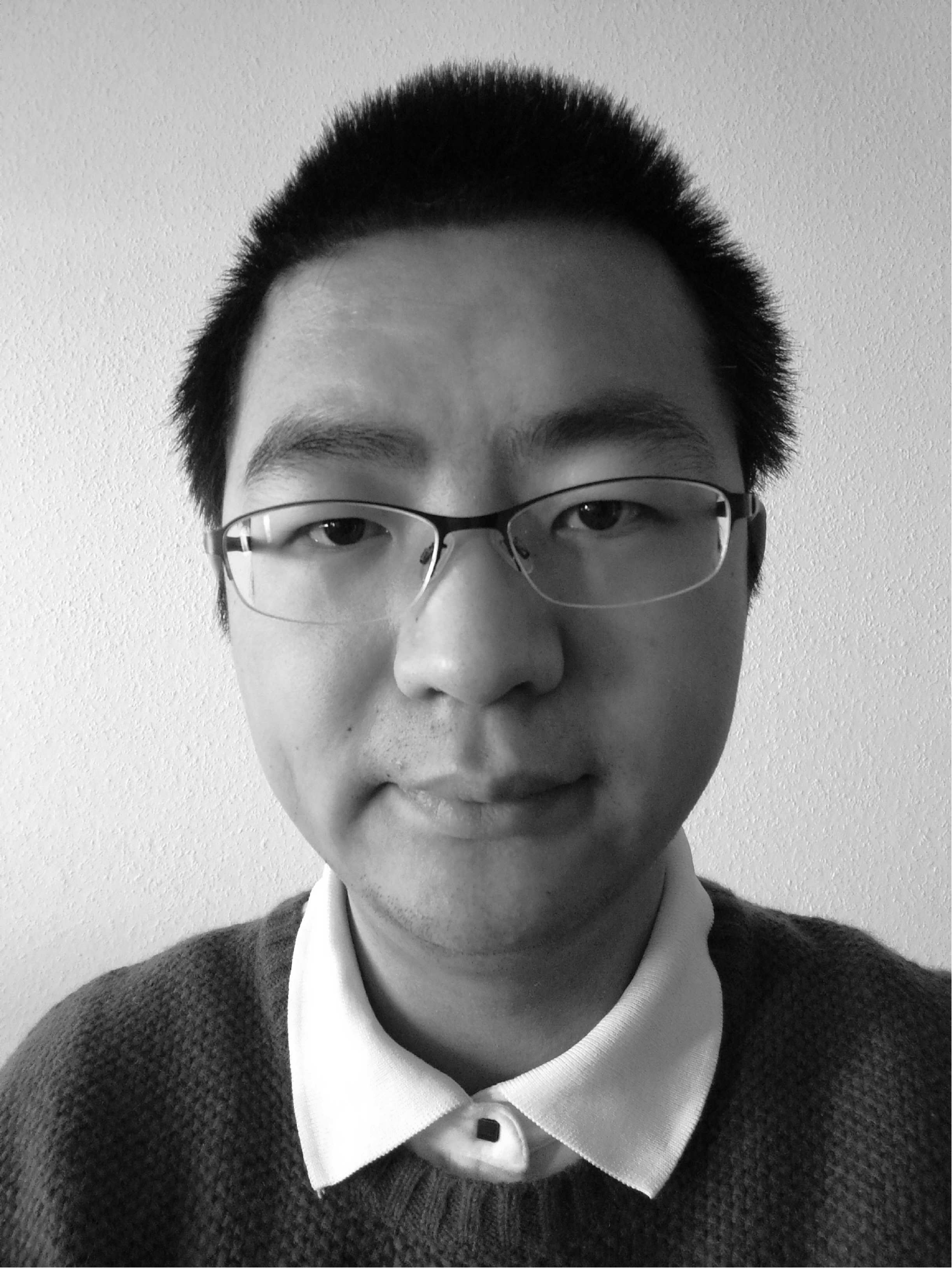}}]{Yexin Yan}
received the Dipl.-Ing. (M.Sc.) in Electrical Engineering from Technische Universität Dresden, Germany, in 2016. He is currently pursuing the Ph.D. at the Chair of Highly-Parallel VLSI-Systems and Neuromorphic Circuits at Technische Universität Dresden. His research interests include hardware-software co-design for applications of brain-inspired algorithms on neuromorphic systems.
\end{IEEEbiography}
\begin{IEEEbiography}[{\includegraphics[width=1in,height=1.25in,clip,keepaspectratio]{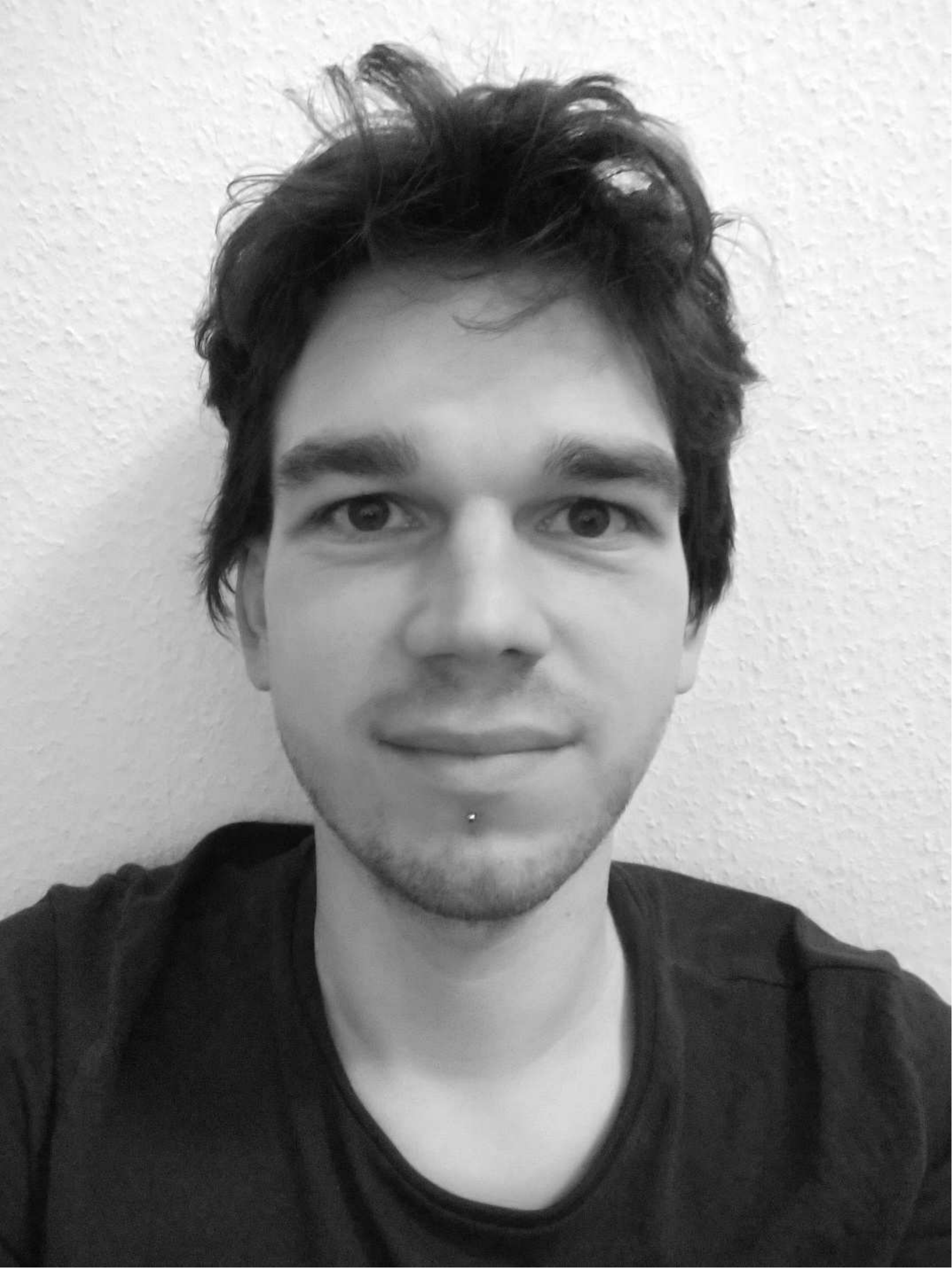}}]{David Kappel}
received the PhD degree in computer science from the Graz University of Technology in 2018. He is currently a postdoctoral researcher with the TU Dresden and the University of Göttingen. His research interest focuses on models for synaptic plasticity, neural dynamics, Bayesian inference and hierarchical learning networks.
\end{IEEEbiography}
\begin{IEEEbiography}[{\includegraphics[width=1in,height=1.25in,clip,keepaspectratio]{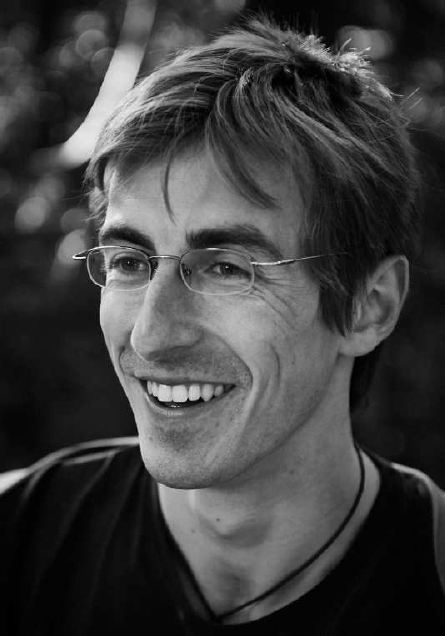}}]{Felix Neumärker}
received the Dipl.-Ing. (M.Sc.) in Electrical Engineering from Technische Universität Dresden, Germany, in 2015.
He is currently working as research associate with the Chair of Highly-Parallel VLSI-Systems and Neuromorphic Circuits at Technische Universität Dresden.
His research interests include software and circuit design and for MPSoCs with special focus on neuromorphic computing.
\end{IEEEbiography}
\begin{IEEEbiography}[{\includegraphics[width=1in,height=1.25in,clip,keepaspectratio]{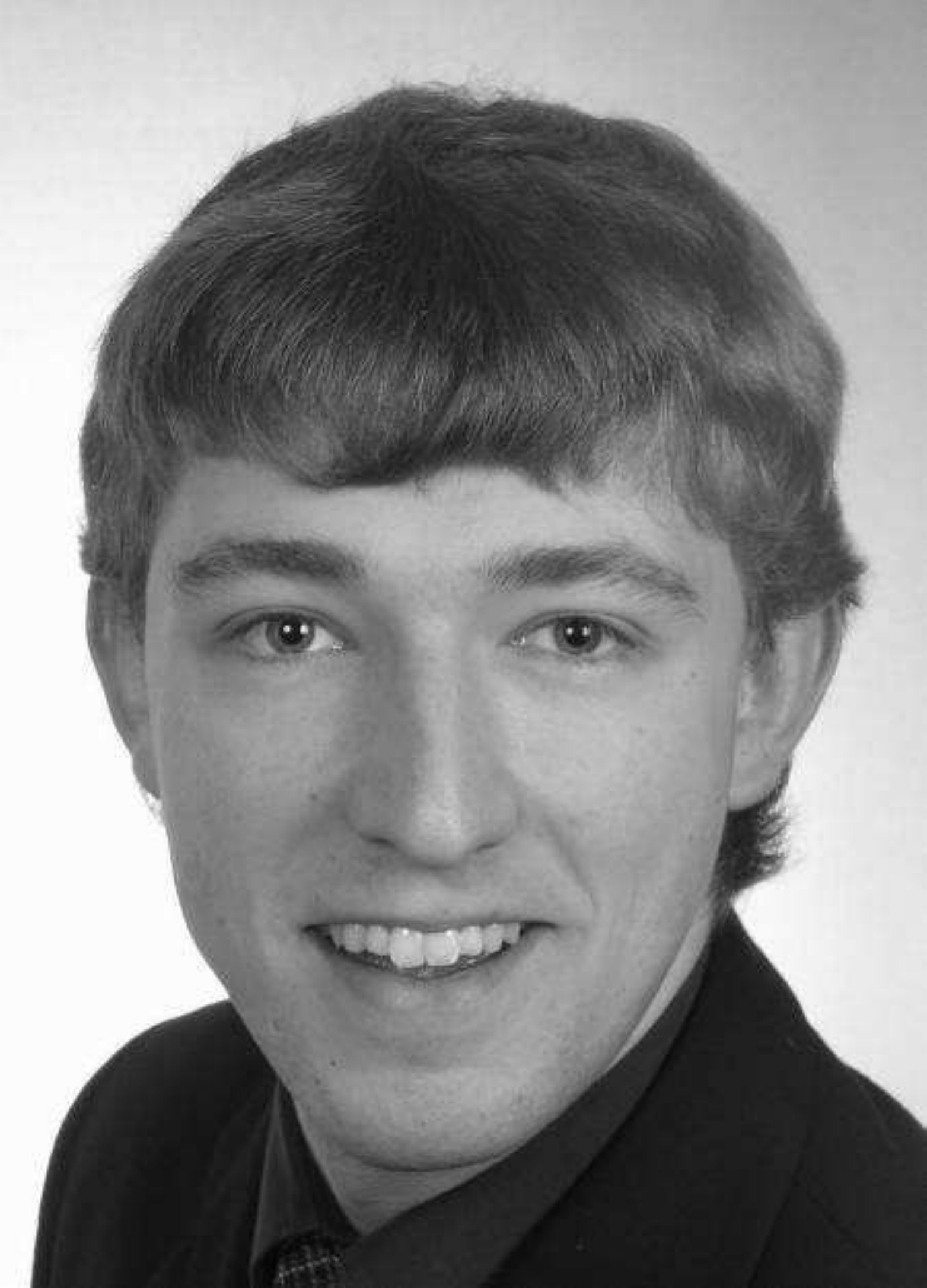}}]{Johannes Partzsch}
obtained his M.Sc.\  in Electrical Engineering in 2007 and his PhD in 2014, both from Technische Universität Dresden. He is currently a Research Group Leader at the Chair of Highly-Parallel VLSI-Systems and Neuromorphic Circuits, Technische Universität Dresden, Germany. His research interests include neuromorphic systems design, topological analysis of neural networks and technical application of bio-inspired systems. He is author or co-author of more than 45 publications.
\end{IEEEbiography}
\begin{IEEEbiography}[{\includegraphics[width=1in,height=1.25in,clip,keepaspectratio]{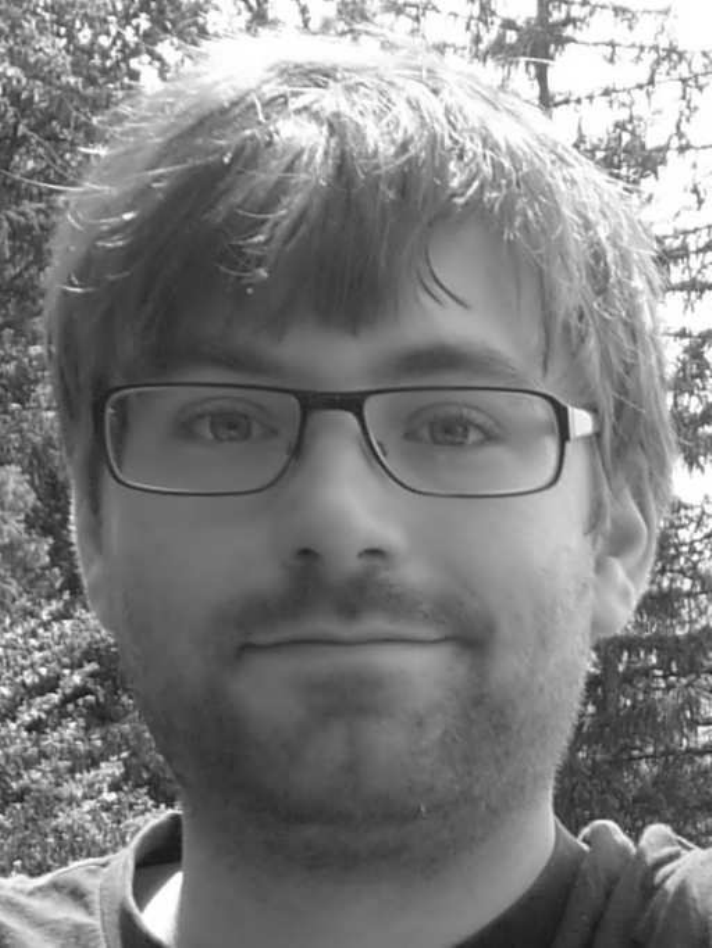}}]{Bernhard Vogginger}
received the diploma in physics from the University of Heidelberg, Heidelberg, Germany, in 2010. Currently, he is a research associate at the Chair of Highly-Parallel VLSI-Systems and Neuromorphic Circuits at Technische Universität Dresden, Germany, where he is pursuing a PhD under the supervision of Prof. Christian Mayr. His research interests include neuromorphic engineering, neural computation and deep learning.
\end{IEEEbiography}
\begin{IEEEbiography}[{\includegraphics[width=1in,height=1.25in,clip,keepaspectratio]{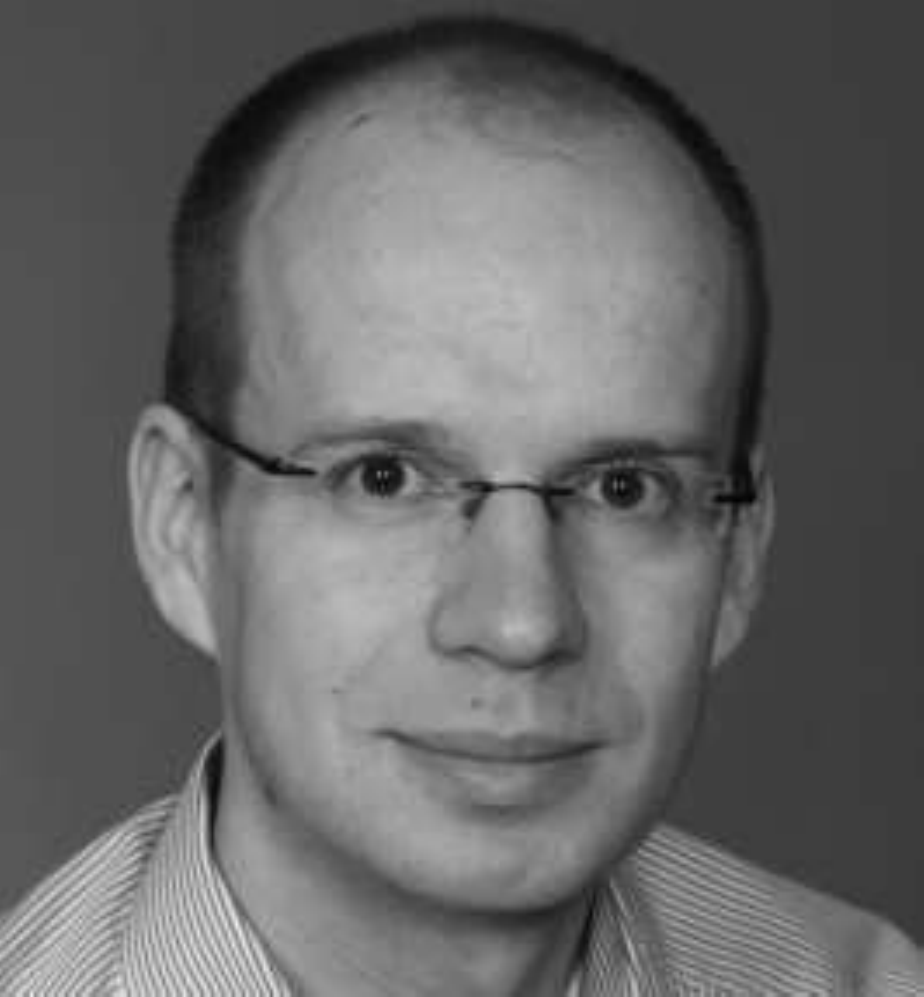}}]{Sebastian Höppner}
is a Research Group Leader and Lecturer with the Chair of Highly-Parallel VLSI-Systems and Neuromorphic Circuits. He received the Dipl.-Ing. (M.Sc.) in Electrical Engineering in 2008 and his Ph.D. in 2013 (received Barkhausen Award), both Technische Universität Dresden, Germany. His research interests include circuits for low-power systems-on-chip in advanced technology nodes, with special focus on clocking, data transmission and power management. He has experience in designing full-custom circuits for multi-processor systems-on-chip (MPSoCs), like ADPLLs, register files and high-speed on-chip and off-chip links, in academic and industrial research projects. He has been managing the full-custom circuit design and SoC integration for more than 12 MPSoC chips in 65nm, 28nm and 22nm CMOS technology. Currently he leads the chip design of the SpiNNaker2 neuromorphic computing system within the Human Brain Project(HBP). He is author or co-author of more than 56 publications and 10 patents (5 issued, 5 pending) in the above fields.
\end{IEEEbiography}
\begin{IEEEbiography}[{\includegraphics[width=1in,height=1.25in,clip,keepaspectratio]{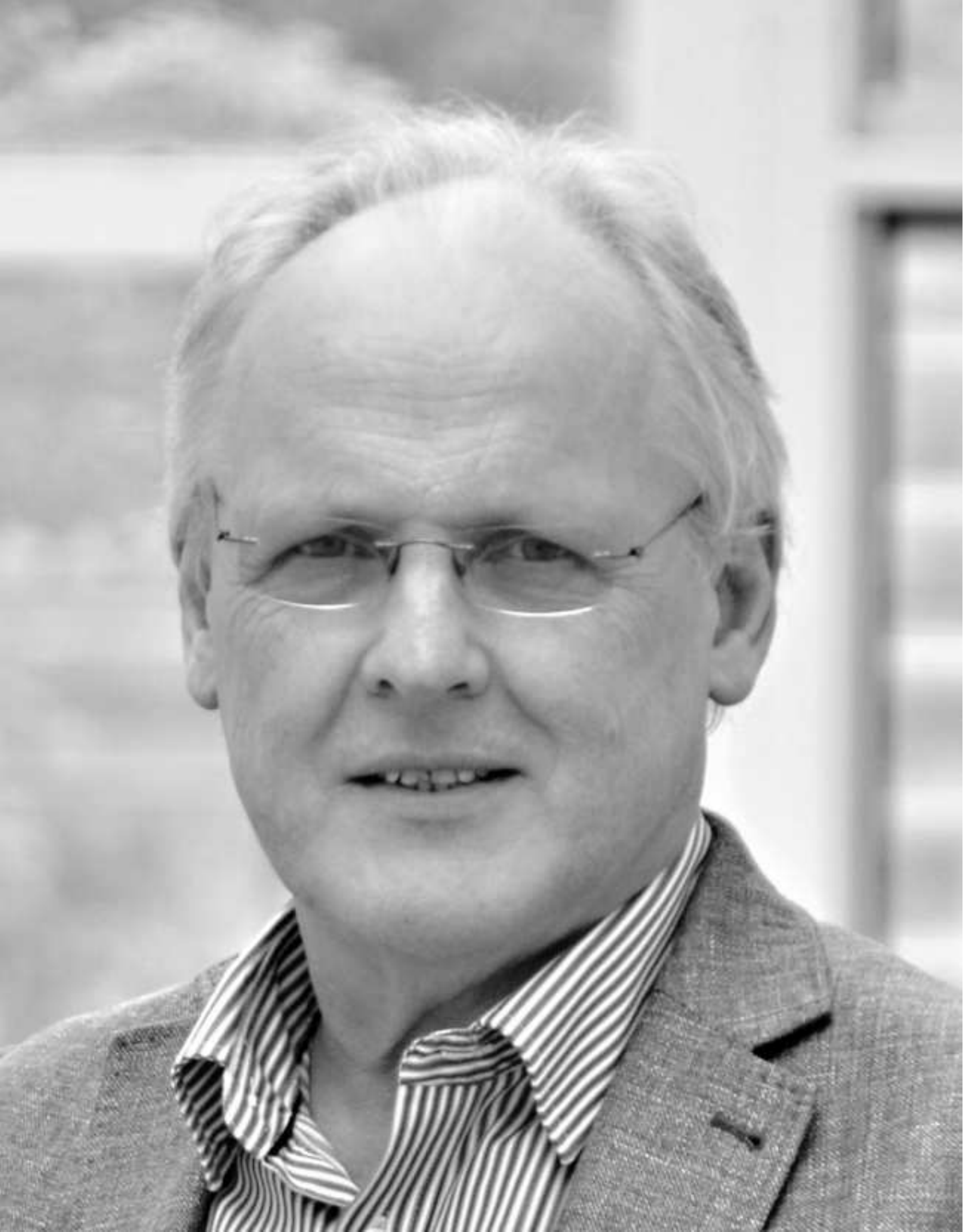}}]{Steve Furber}
CBE FRS FREng is ICL Professor of Computer Engineering in the School of Computer Science at the University of Manchester, UK. After completing a BA in mathematics and a PhD in aerodynamics at the University of Cambridge, UK, he spent the 1980s at Acorn Computers, where he was a principal designer of the BBC Microcomputer and the ARM 32-bit RISC microprocessor. Over 120 billion variants of the ARM processor have since been manufactured, powering much of the world's mobile and embedded computing. He moved to the ICL Chair at Manchester in 1990 where he leads research into asynchronous and low-power systems and, more recently, neural systems engineering, where the SpiNNaker project is delivering a computer incorporating a million ARM processors optimised for brain modelling applications.
\end{IEEEbiography}
\begin{IEEEbiography}[{\includegraphics[width=1in,height=1.25in,clip,keepaspectratio]{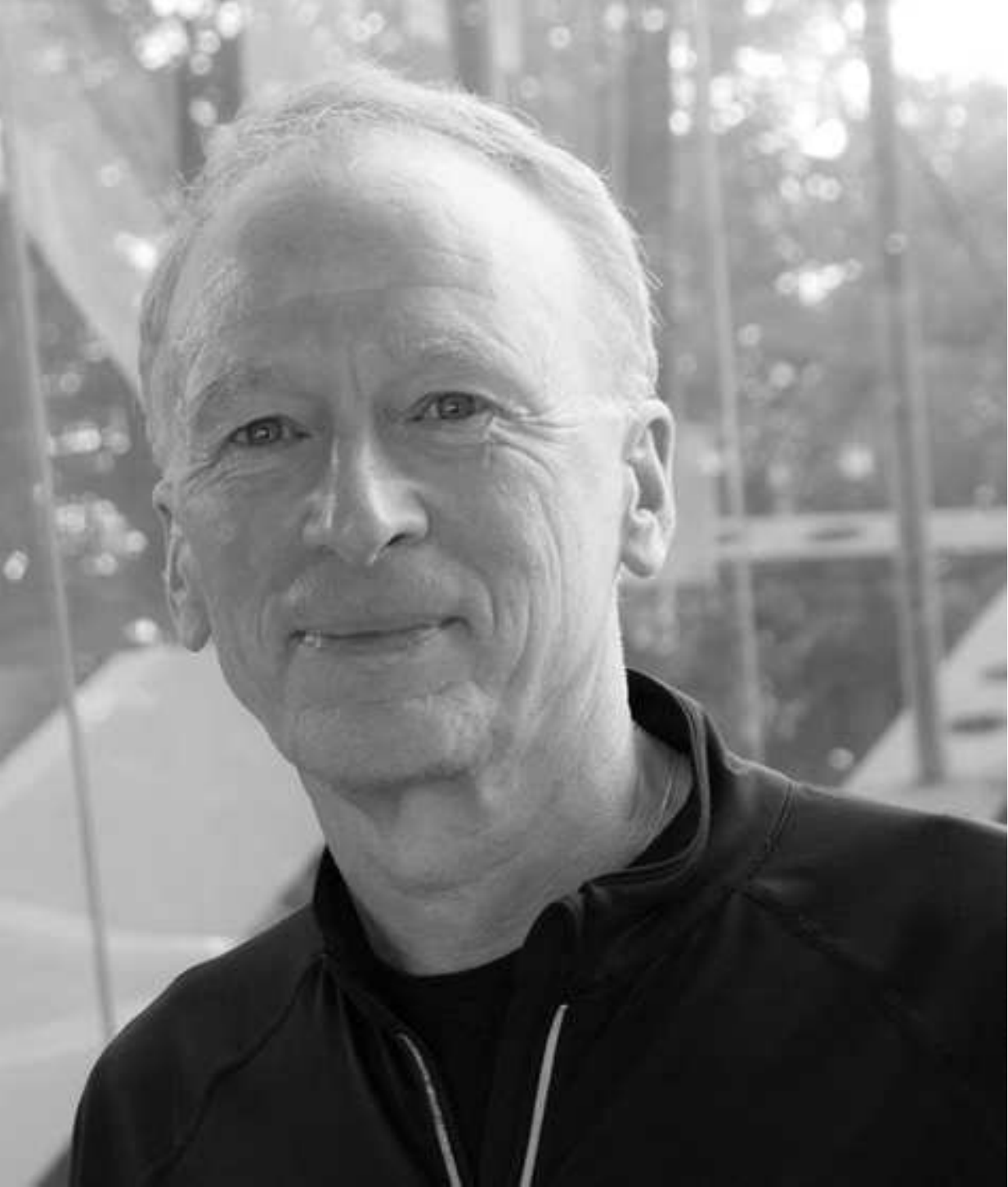}}]{Wolfgang Maass}
received a Phd in Mathematics in 1974, and the Habilitation for Mathematics in 1978 from the Ludwig-Maximilians-Universitaet in Munich. From 1979 to 1984 he was a Postdoc researcher at MIT, the University of Chicago, and the University of California at Berkeley, funded by a Heisenberg-Fellowship of the Deutsche Forschungsgemeinschaft. From 1982 to 1986 he was Associate Professor and from 1986 to 1993 Professor of Computer Science at the University of Illinois in Chicago. Since 1991 he is Professor of Computer Science at the Graz University of Technology in Austria, where he founded the Institut fuer Grundlagen der Informationsverarbeitung (Institute of Theoretical Computer Science). He has authored and co-authored over 240 publications. His research interest is computation and learning in networks of spiking neurons, including implementations in neuromorphic chips.

From 2008 - 2012 he served on the Board of Governors of the International Neural Network Society, and since 2013 he is a member of the Academia Europaea. In 2018 he served as Co-Organizer for the Special Semester "The Brain and Computation" at the Simons Institute, University of California at Berkeley.  
\end{IEEEbiography}
\begin{IEEEbiography}[{\includegraphics[width=1in,height=1.25in,clip,keepaspectratio]{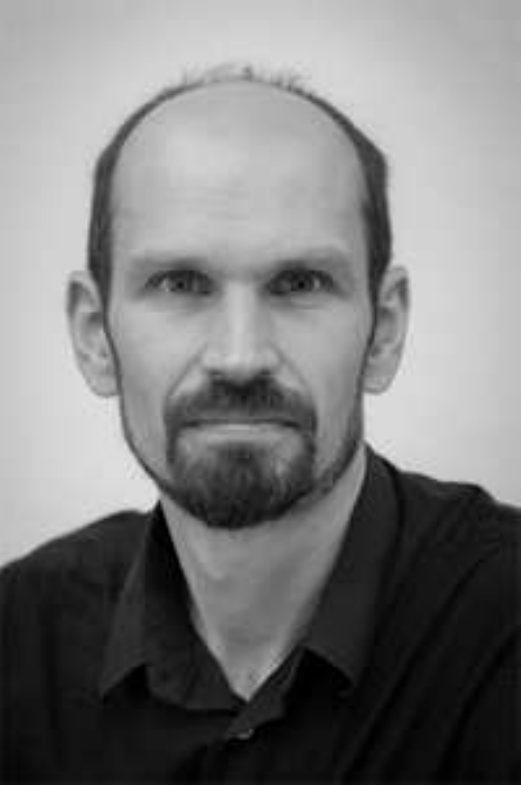}}]{Robert Legenstein}
received the M.Sc. degree from Graz University of Technology, Graz, Austria, in 1999, and the Ph.D. degree in computer science from Graz University of Technology, Graz, Austria, in 2002. He is currently an Associate Professor at the Department of Computer Science, Graz University of Technology and the head of the Institute for Theoretical Computer Science.

Dr. Legenstein has served as associate editor of IEEE Transactions on Neural Networks and Learning Systems (2012-2016) and he was several times in the programme committee for Advances in Neural Information Processing Systems. His primary research interests are learning algorithms in models for biological networks of neurons and neuromorphic hardware, probabilistic neural computation, novel brain-inspired architectures for computation and learning, and memristor-based computing concepts. Dr. Legenstein has established links between biological synaptic plasticity rules and several well-established statistical learning methods such as supervised learning, independent component analysis, and reinforcement learning.
\end{IEEEbiography}
\begin{IEEEbiography}[{\includegraphics[width=1in,height=1.25in,clip,keepaspectratio]{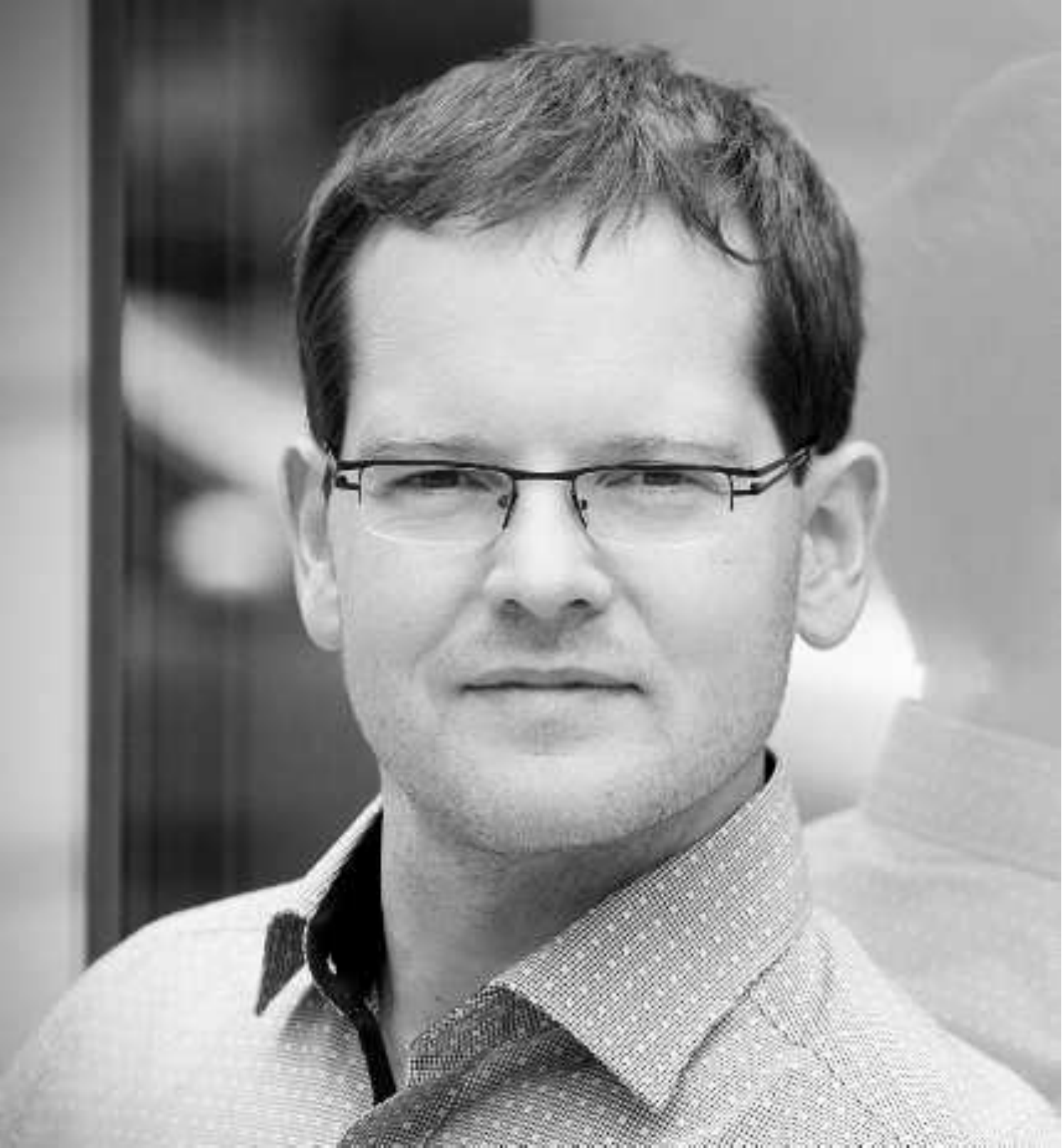}}]{Christian Mayr}
is a Professor of Electrical Engineering at TU Dresden. He received the Dipl.-Ing. (M.Sc.) in Electrical Engineering in 2003, his PhD in 2008 and Habilitation in 2012, all three from Technische Universität Dresden, Germany.
From 2003 to 2013, he has been with Technische Universität Dresden, with a secondment to Infineon (2004-2006). From 2013 to 2015, he did a Postdoc at the Institute of Neuroinformatics, University of Zurich and ETH Zurich, Switzerland. Since 2015, he is head of the Chair of Highly-Parallel VLSI-Systems and Neuromorphic Circuits at Technische Universität Dresden. His research interests include bio-inspired circuits, brain-machine interfaces, AD converters and general mixed-signal VLSI-design. He is author/co-author of over 80 publications and holds 4 patents. He has acted as editor/reviewer for various IEEE and Elsevier journals. His work has received several awards.
\end{IEEEbiography}

\clearpage


\end{document}